%% file: main.tex
\documentclass{article}

\usepackage[preprint]{neurips_2026}

\usepackage[utf8]{inputenc} 
\usepackage[T1]{fontenc}    
\usepackage{hyperref}       
\usepackage{url}            
\usepackage{booktabs}       
\usepackage{amsfonts}       
\usepackage{nicefrac}       
\usepackage{microtype}      

\usepackage{amsmath}
\usepackage{graphicx}
\usepackage{enumitem}
\usepackage{multirow}
\usepackage[dvipsnames]{xcolor}
\usepackage{subcaption}
\usepackage[most]{tcolorbox}

\newtcolorbox{promptbox}[1]{
  colback=gray!7,
  colframe=gray!50,
  fonttitle=\bfseries\footnotesize,
  coltitle=black,
  colbacktitle=gray!20,
  boxrule=0.4pt,
  arc=2pt,
  left=5pt,right=5pt,top=4pt,bottom=4pt,
  title={#1},
  fontupper=\footnotesize\ttfamily,
  breakable,
  enhanced,
}

\iffalse

\newcommand{\nsection}[1]{\section{#1}}
\newcommand{\nsubsection}[1]{\subsection{#1}}

\else

\newcommand{\nsection}[1]{\vspace{-0.1cm}\section{#1}\vspace{-0.1cm}}
\newcommand{\nsubsection}[1]{\vspace{-0.15cm}\subsection{#1}\vspace{-0.1cm}}

\fi

\title{3D Primitives are a Spatial Language for VLMs}

\author{%
  Junze Liu \quad Kun Qian \quad Florian Dubost \quad Kai Zhong \\[4pt]
  {\bfseries Arvind Srinivasan \quad Nan Chen \quad Anping Wang \quad Sam Zhang} \\[4pt]
  {\bfseries Alejandro Mottini \quad Qingjun Cui \quad Tian Wang} \\[8pt]
  \mdseries\normalsize Unity Technologies \\
}

\begin{document}

\maketitle

\begin{abstract}
Vision-language models (VLMs) exhibit a striking paradox: they can generate executable code that reconstructs a 3D scene from geometric primitives with correct object counts, classes, and approximate positions, yet the same models fail at simpler spatial tasks when directly answering natural-language questions on the same image. We show that 3D geometric primitives (cubes, spheres, cylinders, expressed in executable code) serve as a powerful intermediate representation for spatial understanding, and we exploit this insight through three contributions. 
First, we introduce \textbf{\textsc{SpatialBabel}}, a benchmark that evaluates fourteen VLMs on primitive-based 3D scene reconstruction across six \emph{scene-code languages} (programming languages and declarative formats for expressing 3D primitive scenes), revealing that the same model's object-detection F1 can vary by up to $5.7\times$ across scene-code languages. 
Second, we propose \textbf{Code-CoT} (Code Chain-of-Thought), a training-free inference strategy that routes spatial reasoning through primitive-based code generation. Code-CoT improves the SpatialBabel-QA-Score by up to $+6.4$\% on primitive scenes and lifts real-photo CV-Bench-3D accuracy by $+5.0$\% for VLMs with strong coding capabilities. 
Third, we propose \textbf{S$^{3}$-FT} (Self-Supervised Spatial Fine-Tuning), which self-supervisedly distills primitive spatial knowledge into general visual reasoning by parsing the model's own primitive-reconstruction code (generated in Three.js) into structured annotations and fine-tuning on the result, requiring \emph{no human labels and no teacher model}. Training on primitive images alone, S$^3$-FT improves Qwen3-VL-8B by $+4.6$ to $+8.6$\% on SpatialBabel-Primitive-QA, $+9.7$\% on CV-Bench-2D, and $+17$\% on HallusionBench; the recipe transfers across model families. 
These results establish geometric primitives in executable code as both a diagnostic tool for VLM spatial understanding and a transferable spatial vocabulary. 
We will release the benchmark, training data, checkpoints, and evaluation toolkit upon publication.
\end{abstract}

\begin{figure}[t]
\centering
\includegraphics[width=1.0\linewidth]{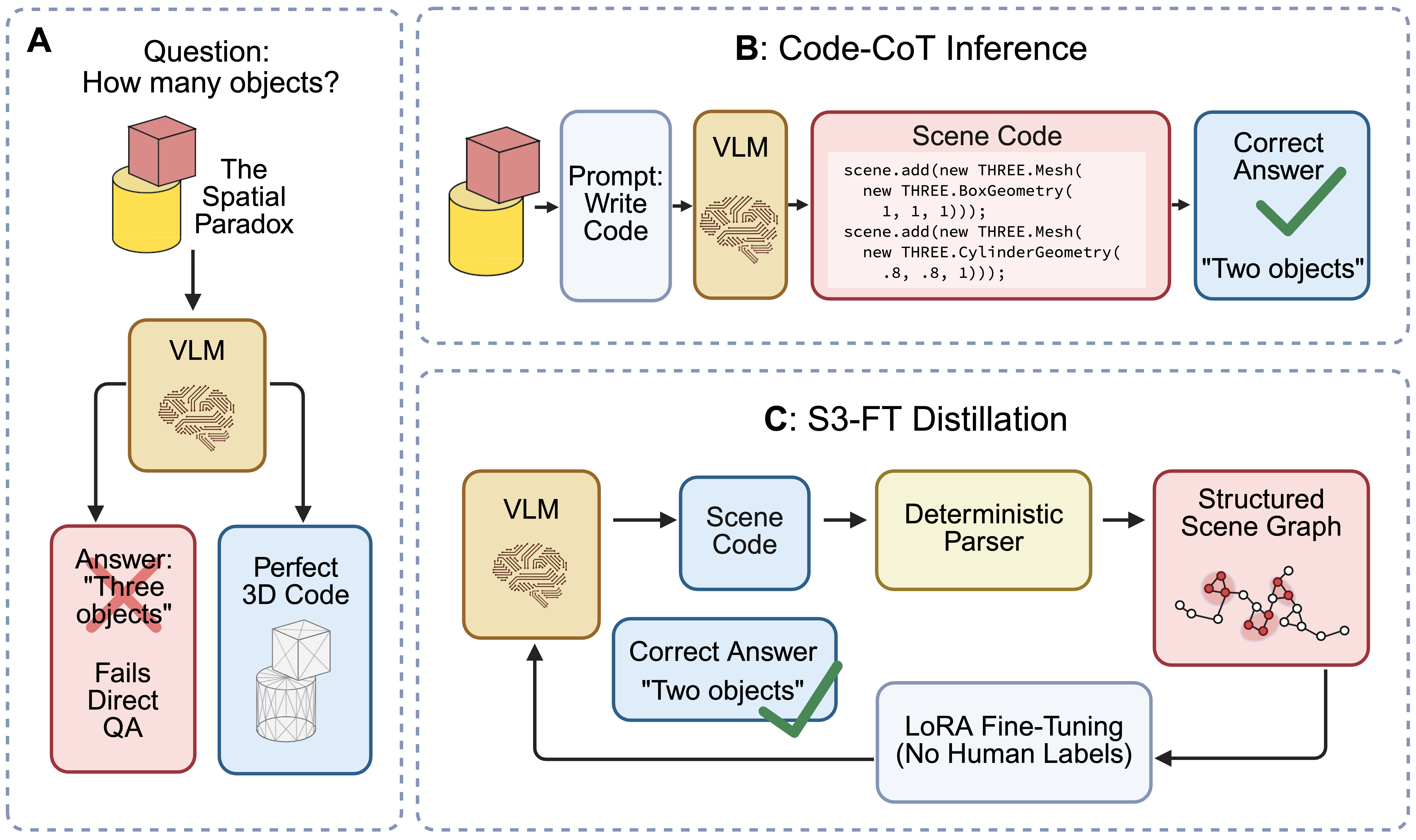}
\caption{Geometric primitives as a spatial language for VLMs: the paradox and two approaches to exploit it. 
\textbf{(A)~The paradox:} A VLM produces a faithful 3D primitive-reconstruction of the input image in code, yet fails simple spatial questions on the same scene
\textbf{(B)~Code-CoT} (Section~\ref{sec:code_cot}) closes the gap \emph{at inference time}: the model first reconstructs the scene as primitive-based executable code, then reasons over its own structured output. 
\textbf{(C)~S$^{3}$-FT} (Section~\ref{sec:s3ft}) closes the gap \emph{at training time}: the model's own primitive-reconstruction outputs are deterministically parsed into structured spatial annotations, which serve as self-supervised labels for LoRA fine-tuning, requiring no human labels. Training on primitives alone yields gains on real-photo benchmarks.}
\label{fig:teaser}
\end{figure}

\input{sections/intro}
\input{sections/related}
\input{sections/benchmark}
\input{sections/analysis}
\input{sections/code_cot}
\input{sections/s3ft}
\input{sections/limitations}

\bibliographystyle{plainnat}
\bibliography{references}

\newpage
\appendix
\input{sections/appendix_extras}
\input{sections/appendix_reproducibility}

\end{document}

%% file: sections/intro.tex
\nsection{Introduction}
\label{sec:intro}

Vision-language models (VLMs) possess more spatial knowledge than their standard outputs reveal. Show a VLM a scene of geometric primitives (a cube and a cylinder) and ask ``how many objects?''; it may hallucinate, answering ``three'' or ``four.'' Yet ask the same model to write scene-reconstruction code (e.g., Three.js) from the same image, and it produces a program that places the correct number of primitives at roughly the correct 3D coordinates (Figure~\ref{fig:teaser}A). The same model, given the same pixels, contradicts itself. Across both proprietary APIs and open-weights models, VLMs generate compilable, spatially faithful 3D primitive-reconstruction programs while simultaneously failing to count, localize, or compose spatial relations when directly answering natural-language questions.

Geometric primitives are uniquely suited for diagnosing this gap: procedurally generated scenes have zero label noise, a no-occlusion constraint removes visibility ambiguity, and simple shapes eliminate object-recognition difficulty, so any failure on these scenes is purely a failure of spatial reasoning. Primitive-reconstruction code is also deterministically parseable into structured scene graphs, enabling automatic evaluation without human annotation.

This paradox points to a \emph{task-routing failure}: scene-code generation is a strict superset of direct QA, as every visible object must be classified, sized, positioned, and named in a single coherent program. So, a model that emits accurate scene code demonstrably possesses the spatial understanding needed to answer the simpler question, yet fails to transfer it to natural-language QA. 

We propose a methodology to test this hypothesis, quantify the phenomenon, and close the gap. Our investigation yields four progressive findings:

\paragraph{Finding 1: A multi-scene-code-language benchmark surfaces the routing failure (Section~\ref{sec:benchmark}).} We introduce \textsc{SpatialBabel}, a benchmark that evaluates VLM spatial understanding through 3D scene reconstruction across six scene-code languages (programming languages and declarative formats for expressing 3D primitive scenes) and a complementary spatial-QA task. 
Reconstruction quality varies sharply with the scene-code language, and each model exhibits a distinct pattern.

\paragraph{Finding 2: Primitive-based Code-CoT is a useful inference scaffold for capable VLMs (Section~\ref{sec:code_cot}).} We introduce Code-CoT (Code Chain-of-Thought): the model first reconstructs the scene as primitive-based executable code, then answers spatial questions conditioned on its own reconstruction. Code-CoT lifts accuracy of the spatial-QA task derived from primitive scenes for mid-tier frontier models and lifts real-photo CV-Bench-3D~\cite{cvbench2024} accuracy for VLMs with strong coding capability.

\paragraph{Finding 3: 3D geometric primitives serve as a compositional spatial vocabulary for VLMs (Section~\ref{sec:code_cot:why_primitives}).} The Code-CoT lifts on primitive scenes and CV-Bench-3D suggest that VLM code training has implicitly tokenized 3D space through geometric primitive shapes. Unlike coordinate-based scaffolds, 
3D primitives encode shape, scale, and orientation in a single structured representation that aligns with how VLMs are trained to write 3D scene code.
\vspace{-2mm}

\paragraph{Finding 4: Self-supervised fine-tuning on primitive scenes distills spatial knowledge to general vision (Section~\ref{sec:s3ft}).} 
S$^3$-FT (Self-Supervised Spatial Fine-Tuning) parses each VLM's own Three.js primitive-reconstruction code into structured spatial annotations and fine-tunes on the resulting multi-task mixture, requiring \emph{no human labels and no teacher model}. Training on primitive images alone yields gains on real-photo benchmarks; the recipe transfers across five base VLMs.

We will release the benchmark, evaluation code, S$^3$-FT training datasets, and checkpoint adapters upon publication.

%% file: sections/related.tex
\nsection{Related Work}
\label{sec:related}

\paragraph{Spatial understanding in VLMs.}
Prior work investigates VLM 3D reasoning through targeted training (SpatialVLM~\cite{chen2024spatialvlm}, SpatialRGPT~\cite{cheng2024spatialrgpt}, SpatialLadder~\cite{li2025spatialladder}) or single-image probing (Theory of Space~\cite{zhang2026theoryofspace}); mechanistic analyses identify structural causes of VLM spatial limitations including vision-token-norm suppression of positional encoding~\cite{qi2025beyond}, 1D flattening of 2D features~\cite{alam2026spatial}, attention imbalance~\cite{chen2025spatial}, and rigid linear spatial bindings~\cite{kang2026linear}. \textsc{SpatialBabel} complements these by evaluating through \emph{generative primitive-reconstruction code}, a stricter task than qualitative QA that externalizes the model's spatial representation into a parseable form.

\paragraph{3D scene generation and spatial-reasoning benchmarks.}
Scene-code generation methods (Scene Language~\cite{zhang2025scene}, SceneScript~\cite{avetisyan2024scenescript}, SceneCraft~\cite{hu2024scenecraft}, BlenderGym~\cite{gu2025blendergym}, Design2Code~\cite{si2025design2code}) and spatial-reasoning benchmarks (SpatialBench~\cite{xu2025spatialbench}, VSI-Bench~\cite{yang2025thinking}, 3DSRBench~\cite{ma20253dsrbench}, the spatial slices of MMMU~\cite{yue2024mmmu} / MathVista~\cite{lu2023mathvista}) each fix a single scene-code language or output channel, conflating spatial competence with format familiarity; code-generation benchmarks (HumanEval~\cite{chen2021evaluating}, MBPP~\cite{austin2021program}, SWE-Bench~\cite{jimenez2023swe}) evaluate programming in isolation from spatial reasoning. \textsc{SpatialBabel} disentangles the two by holding spatial content constant while varying the scene-code language across six formats.

\paragraph{Primitives for spatial reasoning.}
Concurrent work by \citet{lu2026thinkingvisualprimitives} also uses ``primitives'' for VLM spatial reasoning, but with a different definition: DeepSeek interleaves \emph{2D points and bounding boxes} (in pixel coordinates) into the chain of thought as referencing markers, while \textsc{SpatialBabel} uses \emph{3D geometric primitives} 
as compositional scene elements with 3D position, rotation, and scale, expressed in executable code. 2D markers describe \emph{where} an object sits in the image plane; 3D primitives additionally encode \emph{what shape} and \emph{how oriented}. 
Primitives play two roles in our work: the diagnostic stimulus of \textsc{SpatialBabel} and the supervisory signal of S$^3$-FT.

%% file: sections/benchmark.tex
\nsection{The SpatialBabel Benchmark}
\label{sec:benchmark}

Existing benchmarks evaluate VLM spatial reasoning through a single output channel — natural-language QA or a fixed scene-code language, conflating genuine spatial misunderstanding with unfamiliarity with that channel. \textsc{SpatialBabel} disentangles the two by combining multi-scene-code 3D reconstruction with spatial QA. 

\begin{figure}[t]
\centering
\includegraphics[width=0.98\linewidth]{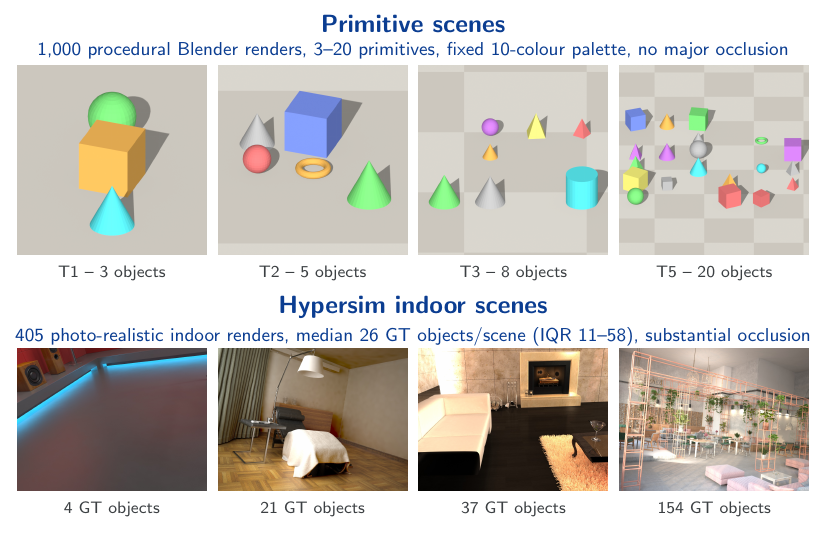}
\caption{\textsc{SpatialBabel}'s two scene domains. \textbf{Top:} four primitive scenes (T1, T2, T3, T5) from the 1{,}000-scene procedural split (fixed 10-color palette, no-major-occlusion placement). \textbf{Bottom:} four Hypersim photo-realistic indoor scenes (median 26 ground-truth objects per scene).}
\label{fig:dataset_overview}
\end{figure}

\nsubsection{Scene domains, tasks, and scene-code languages}
\label{sec:scenes}\label{sec:tasks}\label{sec:languages}

\paragraph{Two scene domains with distinct diagnostic roles.}
The benchmark spans two complementary scene domains, chosen to enable systematic study of primitive-to-complex transfer (Figure~\ref{fig:dataset_overview}). The \emph{primitive} split (1{,}000 scenes) contains procedurally-generated Blender scenes with 3--20 geometric primitives (cubes, spheres, cylinders, cones, tori, pyramids) organized into five complexity tiers (T1 = 3 objects through T5 = 20). Colors are drawn from a fixed 10-color palette, and a no-major-occlusion placement constraint ensures that every object is clearly visible and identifiable. Primitives serve as the \emph{primary diagnostic tool} for three reasons: (i) procedural generation with known object placements provides exact ground truth with zero label noise; (ii) the no-occlusion constraint eliminates visibility ambiguity, so that any spatial failure is purely a reasoning failure rather than a perception one; (iii) primitive-reconstruction code can be deterministically parsed into structured scene graphs, enabling fully automatic evaluation. The \emph{Hypersim} split (405 scenes, 200 evaluated) contains photo-realistic indoor scenes from~\cite{roberts2021hypersim}, from which we extract per-object ground truth from the original 3D meshes. Hypersim serves as the \emph{harder validation domain}: 60 object classes, complex geometries, and substantial inter-object occlusion. 
We evaluate on a representative 200-scene subset stratified across scene complexity levels to balance evaluation cost with statistical coverage. 
All renders are $512 \times 512$ from a single frontal viewpoint.

\vspace{-2mm}
\paragraph{Tasks.}
\textsc{SpatialBabel} exposes two complementary task families derived from the same underlying scenes. The primary diagnostic is \emph{cross-scene-code-language reconstruction} (\textbf{Reconstruct}): given a single-view image, the model must emit executable code in a specified scene-code language that reconstructs the depicted scene. The secondary task family, \emph{programmatic spatial question answering} (\textbf{QA}), provides approximately eight natural-language questions per scene drawn from five categories: localization, relationship, counting, existence, and comparison. Each task can be evaluated on either scene domain, yielding the four settings: SpatialBabel-Primitive-Reconstruct, SpatialBabel-Primitive-QA, SpatialBabel-Hypersim-Reconstruct, and SpatialBabel-Hypersim-QA.

\vspace{-2mm}
\paragraph{Scene-code languages.}
We evaluate reconstruction across six scene-code languages spanning different programming paradigms and varying levels of pre-training prevalence: Three.js (imperative web 3D via JavaScript), Unity C\# (game-engine API), Blender Python (procedural scripting), Open3D Python (mesh-creation API), Canonical JSON (a declarative schema specifying \texttt{position}/\texttt{rotation}/\texttt{scale}/\texttt{class\_name}, introduced by this work), and a Scene Language DSL~\cite{zhang2025scene}. The multi-scene-code design is the key diagnostic lever: a scene-code-agnostic spatial representation should yield roughly invariant performance across scene-code languages, whereas spatial knowledge entangled with a particular code-generation prior should produce model-specific cross-scene-code gaps. Pre-training exposure proxies for each scene-code language are reported in Appendix~\ref{app:exposure_proxies}.

\nsubsection{Evaluation protocol and metrics}
\label{sec:eval_protocol}\label{sec:metrics}

For each (model, scene-code language) pair, the generated code is deterministically parsed into a structured scene representation consisting of geometric primitives---including for Hypersim scenes, where complex objects are approximated by their bounding primitives. Predicted objects are then optimally matched to ground-truth objects via the Hungarian algorithm~\cite{kuhn1955hungarian}, minimizing Euclidean position distance. Each (model, scene-code language) cell is scored on five $[0, 1]$ components: \emph{parse rate} (fraction of outputs producing valid \texttt{Scene} objects), \emph{F1} (harmonic mean of precision and recall over Hungarian-matched objects), \emph{class accuracy} (synonym-aware accuracy on matched pairs), \emph{position fidelity} $= \max(0,\, 1 - \min(d_\text{pos}, 1))$ where $d_\text{pos}$ is the per-scene root-mean-square Euclidean position error normalized by the ground-truth scene's spatial extent (clipped per-scene before averaging), and \emph{scale fidelity} $= \max(0,\, 1 - \min(d_\text{scale}, 1))$ where $d_\text{scale}$ is the per-object root-mean-square error between predicted and ground-truth $(s_x, s_y, s_z)$ scale vectors. These five components are aggregated into the \emph{SpatialBabel-Reconstruct-Score}:
\begin{equation}
  \text{Reconstruct Score} = \tfrac{1}{5}\bigl(\text{Parse Rate} + \text{F1} + \text{Class Accuracy} + \text{Position Fidelity} + \text{Scale Fidelity}\bigr) \in [0, 1].
\label{eq:reconstruct_score}
\end{equation}

We evaluate 14 VLMs (8 proprietary APIs and 6 open-weights models) under two prompting modes (Baseline and natural-language chain-of-thought). Per-scene-code Reconstruct Scores on the primitive split (SpatialBabel-Primitive-Reconstruct) are reported in Table~\ref{tab:main_results}, with per-model sample counts and prompting details in Appendix~\ref{app:models_prompting}.

%% file: sections/analysis.tex

\nsubsection{Analysis}
\label{sec:analysis}\label{sec:main_results}\label{sec:finding_reconstruct_vs_qa}\label{sec:finding_crosslang}\label{sec:finding_hypersim}\label{sec:from_diagnosis}

\begin{table}[t]
\centering
\caption{Per-scene-code SpatialBabel-Reconstruct-Score on primitive scenes (SpatialBabel-Primitive-Reconstruct
) for fourteen VLMs under \emph{Direct} prompting. Each cell is a $[0,1]$ aggregate of parse rate, F1, class accuracy, and position/scale fidelity (Section~\ref{sec:metrics}). \textbf{Bold}: best per row. \colorbox{red!15}{Red}: lowest per row. Coding languages are ordered by approximate pre-training exposure (most $\rightarrow$ least; proxies in Appendix~\ref{app:exposure_proxies}). $\Delta$: max--min Reconstruct-Score gap within a row. Results under natural-language chain-of-thought prompting are reported in the per-component breakdown (Appendix Table~\ref{tab:sbs_components}).}
\label{tab:main_results}
\small
\setlength{\tabcolsep}{4pt}
\begin{tabular}{l cccccc c}
\toprule
Model & Three.js & Unity & Blen.Py & Open3D & JSON & DSL & $\Delta$ \\
\midrule
\multicolumn{8}{l}{\emph{Proprietary APIs}} \\
Claude Opus 4.7  & \textbf{.792} & .679 & .739 & .702 & .734 & \colorbox{red!15}{.615} & .177 \\
Claude Sonnet 4.6 & .693 & .681 & .726 & .713 & \textbf{.733} & \colorbox{red!15}{.526} & .207 \\
GPT-5            & .646 & \colorbox{red!15}{.473} & .630 & .596 & \textbf{.677} & .628 & .205 \\
GPT-4o           & \textbf{.730} & .672 & .675 & .705 & .719 & \colorbox{red!15}{.658} & .071 \\
Gemini-2.5-Pro   & \textbf{.732} & \colorbox{red!15}{.481} & .594 & .689 & .729 & .594 & .250 \\
Gemini-2.5-Flash & .672 & .586 & .657 & .678 & \textbf{.772} & \colorbox{red!15}{.474} & .298 \\
Gemini-3-Flash   & .684 & \colorbox{red!15}{.543} & .652 & .723 & \textbf{.793} & .680 & .249 \\
Gemini-3-Pro     & \textbf{.834} & .680 & .716 & .712 & .812 & \colorbox{red!15}{.575} & .259 \\
\midrule
\multicolumn{8}{l}{\emph{Open weights}} \\
Qwen3-VL-8B      & .640 & .651 & \colorbox{red!15}{.514} & .617 & \textbf{.676} & .629 & .162 \\
Qwen2.5-VL-32B   & .686 & .656 & \colorbox{red!15}{.471} & \textbf{.707} & .603 & .509 & .236 \\
Qwen2.5-VL-7B    & \textbf{.599} & .376 & .320 & .599 & .566 & \colorbox{red!15}{.315} & .284 \\
Qwen2.5-VL-3B    & .546 & .618 & \textbf{.667} & .357 & .550 & \colorbox{red!15}{.321} & .346 \\
LLaVA-OV-7B      & .640 & \textbf{.668} & .648 & .596 & \colorbox{red!15}{.421} & .652 & .248 \\
InternVL3-8B     & \colorbox{red!15}{.541} & .563 & \textbf{.649} & .576 & .591 & .641 & .108 \\
\bottomrule
\end{tabular}
\end{table}

Table~\ref{tab:main_results} reports per-scene-code Reconstruct Scores for fourteen VLMs on primitive scenes. \emph{No model performs consistently across scene-code languages}: max--min gaps within a single model range from $0.07$ (GPT-4o) to $0.35$ (Qwen2.5-VL-3B), and the underlying F1 component spans up to $5.7\times$ on the same images. The best/worst language is itself model-specific, 
so spatial knowledge is entangled with each model's scene-code-specific code-generation prior. The pattern persists from primitive to photo-realistic scenes (mean Reconstruct Score drops $0.11$--$0.27$ on Hypersim, but per-model rankings remain largely stable; Appendix~\ref{app:hypersim_table}).

Reconstruction skill does not predict QA performance. Across the nine VLMs evaluated on both metrics, mean Reconstruct-Score F1 spans $0.44$--$0.94$, while relationship-category QA accuracy clusters within a $2.5$-pp band ($35.7$--$38.2$\%; Pearson $r=0.12$, Spearman $\rho=0.10$; per-model breakdown in Appendix Table~\ref{tab:sbs_qa_corr}). 
The per-model rankings motivate the two interventions that follow: a per-model best/worst scaffold for inference-time Code-CoT (Section~\ref{sec:code_cot}) and a structured pseudo-ground-truth signal for self-supervised training (Section~\ref{sec:s3ft}).

%% file: sections/code_cot.tex
\providecommand{\deltaup}[1]{{\color{ForestGreen}\scriptsize$\,(+#1)$}}
\providecommand{\deltadn}[1]{{\color{BrickRed}\scriptsize$\,(-#1)$}}
\providecommand{\deltazero}[1]{{\color{gray}\scriptsize$\,(\pm#1)$}}

\nsection{Code-CoT: Primitive Reconstruction as a Generic Spatial Reasoning Tool}
\label{sec:code_cot}

We exploit the model's spatial knowledge pathway of primitive-reconstruction with Code Chain-of-Thought (Code-CoT), a training-free strategy where the model first emits primitive-based scene code and then answers spatial questions conditioned on it, across three domains: \textsc{SpatialBabel} primitives, Hypersim photo-realistic indoors, and CV-Bench-3D real photos.

\nsubsection{Setup}
\label{sec:code_cot:setup}

We compare four inference modes on the same SpatialBabel-QA prompts: (i) \textbf{direct} (image + question, no reasoning instruction); (ii) \textbf{NL-CoT}, a natural-language chain-of-thought scaffold; (iii) \textbf{best-language Code-CoT}, which reconstructs the scene in the model's primitive-best scene-code language (Table~\ref{tab:main_results}) before answering; (iv) \textbf{worst-language Code-CoT}, same procedure with the primitive-worst language. The NL-CoT prompt, the proprietary-API direct-mode caveat, the localization-category note, and the CoT-symmetric QA evaluator are described in Appendix~\ref{app:cot_evaluator}.

\nsubsection{Results across three domains}
\label{sec:code_cot:results}\label{sec:code_cot:spatialbabel}\label{sec:code_cot:hypersim}\label{sec:code_cot:cvbench3d}

We evaluate Code-CoT on three domains: SpatialBabel-Primitive-QA, SpatialBabel-Hypersim-QA, and CV-Bench-3D~\citep{cvbench2024} (1{,}200 real-photo 3D spatial-relation MCQs). Per-cell tables for the latter two are in Appendices~\ref{app:code_cot_hypersim} and~\ref{app:cvbench_table}.

\begin{table}[t]
\centering\small
\caption{SpatialBabel-QA-Score (\%) under four inference modes on SpatialBabel-Primitive-QA (100 scenes $\times$ 8 programmatic questions = 800 per cell). \texttt{best\_cc} / \texttt{worst\_cc} use each model's primitive-best / primitive-worst scene-code language from Table~\ref{tab:main_results} restricted to the 4-language subset \{Three.js, Canonical JSON, Blender Python, Scene Language DSL\}. Bold = row max. All cells use the CoT-symmetric evaluator (Appendix~\ref{app:cot_evaluator}).}
\label{tab:code_cot_sb}
\begin{tabular}{lccccc}
\toprule
Model & direct & nl\_cot & best\_cc & worst\_cc & $\Delta_\text{best-direct}$ \\
\midrule
\multicolumn{6}{l}{\emph{Proprietary APIs}} \\
Claude Opus 4.7         & 48.8          & \textbf{49.2} & 47.6          & 47.9          & \deltadn{1.1} \\
Claude Sonnet 4.6       & 43.1          & 42.6          & \textbf{46.9} & \textbf{46.9} & \deltaup{3.7} \\
GPT-5                   & 45.1          & 45.0          & 45.6          & \textbf{46.0} & \deltaup{0.5} \\
GPT-4o                  & 43.0          & 41.9          & 45.1          & \textbf{46.9} & \deltaup{2.1} \\
Gemini-2.5-Pro          & 45.5          & \textbf{52.8} & 51.9          & 51.4          & \deltaup{6.4} \\
Gemini-2.5-Flash        & \textbf{48.1} & 46.2          & \textbf{48.1} & \textbf{48.1} & \deltazero{0.0} \\
Gemini-3-Flash          & 47.4          & 49.6          & \textbf{50.1} & 46.6          & \deltaup{2.7} \\
Gemini-3-Pro            & 47.2          & \textbf{47.4} & 46.4          & 43.8          & \deltadn{0.9} \\
\midrule
\multicolumn{6}{l}{\emph{Open weights}} \\
InternVL3-8B            & 37.5          & \textbf{41.4} & 37.9          & 40.5          & \deltaup{0.4} \\
LLaVA-OV-7B             & 40.0          & 37.1          & 42.6          & \textbf{49.4} & \deltaup{2.6} \\
Qwen3-VL-8B             & \textbf{39.1} & 38.6          & 37.5          & 29.8          & \deltadn{1.6} \\
Qwen2.5-VL-7B           & 45.4          & 44.4          & 44.5          & \textbf{46.5} & \deltadn{0.9} \\
Qwen2.5-VL-3B           & 36.4          & \textbf{41.8} & 18.8          & 32.8          & \deltadn{17.6} \\
\bottomrule
\end{tabular}
\end{table}

Table~\ref{tab:code_cot_sb} reports SpatialBabel-Primitive-QA accuracy. Mid-tier frontier models gain the most from primitive-best Code-CoT (Gemini-2.5-Pro $+6.4$\%, Sonnet $+3.7$\%, Gemini-3-Flash $+2.7$\%, GPT-4o $+2.1$\%); ceiling-saturated frontier models (Opus 4.7, GPT-5, Gemini-3-Pro) converge across modes within $\pm 1$\%; weak open-weights coders regress (Qwen-VL family at $-0.9$\% to $-17.6$\%), which can be addressed with S$^3$-FT (Section~\ref{sec:s3ft}) by lifting reconstruction quality across all scene-code languages. 
The cross-domain story is mixed. On Hypersim (Appendix Table~\ref{tab:code_cot_hypersim}), primitive-best Code-CoT is sign-mixed, indicating a domain ceiling once cluttered scenes break the primitive-shape vocabulary. On CV-Bench-3D (Appendix Table~\ref{tab:code_cot_cvbench}), primitive-best Code-CoT lifts the strongest code generators (Sonnet $+5.9$\%, Opus 4.7 $+5.0$\%, GPT-5 $+3.2$\%) while ceiling-saturated frontier Geminis stay within $\pm 1$\% and weak coders regress sharply (Qwen2.5-VL-7B $-11.5$\%, Qwen3-VL-8B $-9.7$\%). Two qualifications recur across all three domains. NL-CoT often matches or exceeds primitive-best Code-CoT on raw accuracy (e.g., Gemini-2.5-Pro NL-CoT $52.8$\% vs.\ best\_cc $51.9$\% on primitive QA); the value of Code-CoT relative to NL-CoT is the cross-language axis it exposes (each row's best\_cc--worst\_cc gap), not headline accuracy alone. And primitive-best struggles on cluttered indoor and real-photo scenes an observation we discuss further in Section~\ref{sec:limitations}.

\nsubsection{Why 3D geometric primitives work: code as spatial tokenization}
\label{sec:code_cot:why_primitives}

Where Code-CoT helps the most (primitive scenes and a subset of CV-Bench-3D) suggests that VLM code training has tokenized 3D space through geometric primitive shapes. Writing Three.js code to place a \texttt{BoxGeometry} at $(2, 0, -3)$ with scale $(1, 2, 1)$ expresses spatial structure in a vocabulary that the language-model backbone has been extensively trained on. Three properties make primitives a useful scaffold beyond natural-language and 2D-coordinate alternatives: they are \emph{compositional} (any scene decomposes into canonical shapes parameterized by position, rotation, scale); they are \emph{richer than only 2D coordinates}; and the primitive approximation preserves the spatial relations that drive QA (above, left-of, in-front-of, closer-than) even when geometry is simplified.

The qualifier is a domain ceiling: when scenes cannot be tokenized into a small primitive vocabulary (cluttered Hypersim indoors, weak coders), the same reconstruction step that helps elsewhere hurts, and NL-CoT, which avoids the primitive commitment, is the safer scaffold. Code-CoT is therefore a primitive-vocabulary scaffold, useful where the vocabulary suffices and a baseline elsewhere.

%% file: sections/s3ft.tex
\providecommand{\deltaup}[1]{{\color{ForestGreen}\scriptsize$\,(+#1)$}}
\providecommand{\deltadn}[1]{{\color{BrickRed}\scriptsize$\,(-#1)$}}
\providecommand{\deltazero}[1]{{\color{gray}\scriptsize$\,(\pm#1)$}}

\nsection{S$^3$-FT: Self-Supervised Distillation of Primitive Spatial Knowledge}
\label{sec:s3ft}

S$^3$-FT targets the weak-coder regime that Code-CoT cannot address (Section~\ref{sec:code_cot}) by distilling primitive spatial knowledge into the same VLM that generated the pseudo-ground-truth, with no human labels and no teacher model.

\nsubsection{Pipeline}
\label{sec:s3ft:pipeline}

\begin{figure}[t]
\centering
\includegraphics[width=1.00\linewidth]{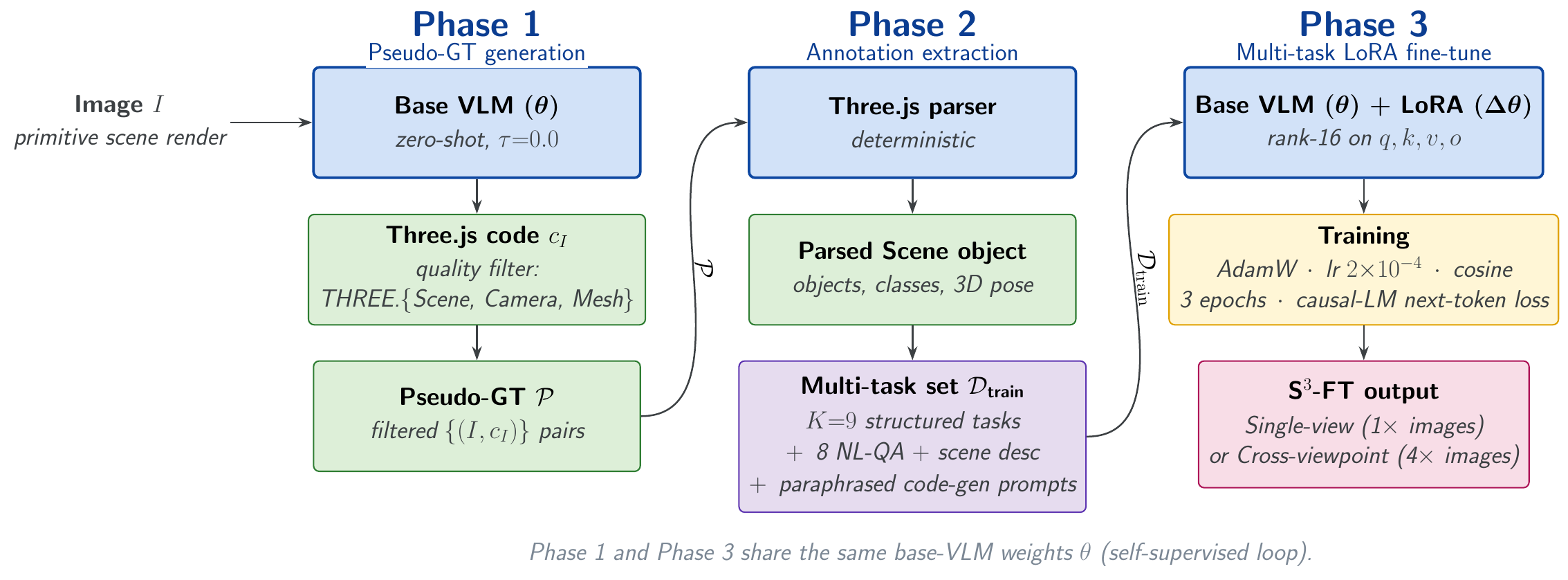}
\caption{The three-phase S$^3$-FT pipeline (Phase 1: pseudo-GT generation; Phase 2: structured annotation extraction; Phase 3: multi-task LoRA fine-tuning).}
\label{fig:s3ft_pipeline}
\end{figure}

From a set of primitive scene images $\mathcal{I}$, S$^3$-FT proceeds in three phases (Figure~\ref{fig:s3ft_pipeline}). \textbf{Phase 1} queries the base VLM zero-shot for a Three.js primitive-reconstruction per image and applies a syntactic quality filter (\texttt{THREE.Scene}/\texttt{Camera}/\texttt{Mesh} present; object count in $[1,50]$; positions in $[-30,30]$) to form a pseudo-ground-truth set $\mathcal{P} = \{(I, c_I)\}$. \textbf{Phase 2} parses each $c_I$ deterministically into a \texttt{Scene} and derives $K{=}9$ structured-task targets (object count, classes, 3D positions, 3D / 2D bounding boxes, spatial relations, depth ordering, scene graph, code generation) plus rule-generated natural-language signals (8 QA pairs, a scene-description paragraph, and two paraphrases of the code-gen prompt per scene); the NL signals are essential, as an ablation that drops them collapses the model's natural-language behaviour (Appendix~\ref{app:s3ft:collapse}). \textbf{Phase 3} fine-tunes the same VLM with a rank-16 LoRA on q/k/v/o projections (cosine schedule, 3 epochs); we call this configuration Single-view S$^3$-FT. Every supervised signal originates from the VLM's own output in $\mathcal{P}$, so the procedure requires no human labels and no teacher model.

\nsubsection{In-distribution results}
\label{sec:s3ft:single-view}

\begin{figure}[t]
\centering
\includegraphics[width=1.00\linewidth]{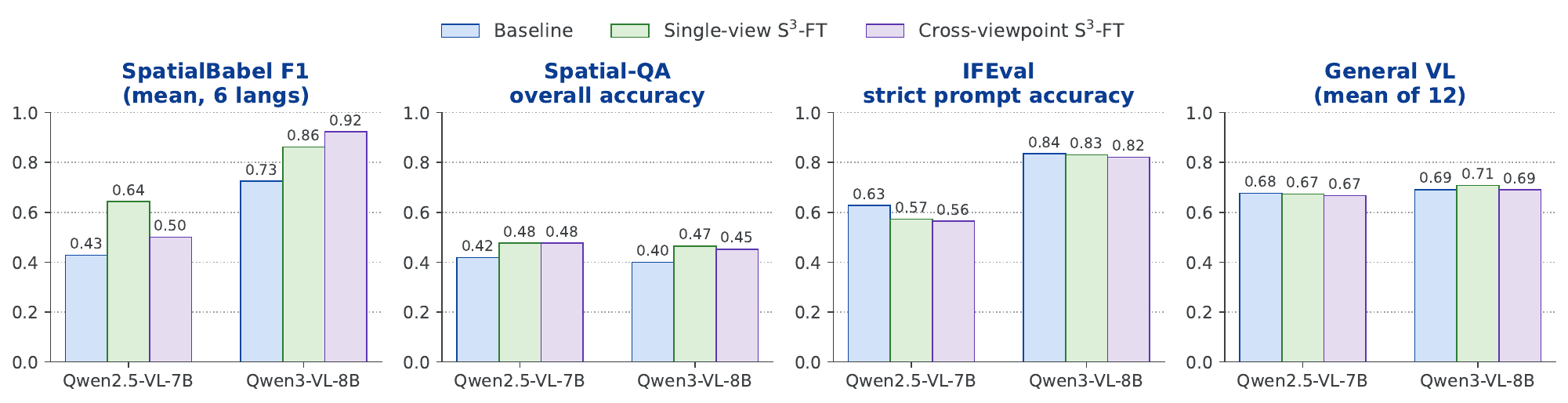}
\caption{S$^3$-FT vs. Baseline on two base VLMs. Panel 1: SpatialBabel-Primitive-Reconstruct mean F1 (in-distribution). Panel 2: SpatialBabel-Primitive-QA accuracy. Panel 3: IFEval. Panel 4: mean across 12 standard VLM benchmarks (out-of-distribution).}
\label{fig:s3ft_summary}
\end{figure}

Figure~\ref{fig:s3ft_summary} summarizes the four-axis evaluation. Mean SpatialBabel-Primitive-Reconstruct F1 rises from $0.43$ to $\mathbf{0.64}$ (Qwen2.5-VL-7B) and from $0.73$ to $\mathbf{0.86}$ (Qwen3-VL-8B) under Single-view S$^3$-FT. On SpatialBabel-Primitive-QA, S$^3$-FT lifts Qwen3-VL-8B uniformly across all four inference modes ($+4.6$ to $+8.6$\%, mean $+7.0$\%; Table~\ref{tab:code_cot_sb_sft}); on Qwen2.5-VL-7B three of four modes improve ($+3.2$\%, $+3.1$\%, $+1.7$\%) but primitive-worst Code-CoT regresses ($-8.2$\%) because same-distribution distillation overwrites that base's already-clean worst-language CoT, leaving the four-mode mean roughly flat. The Qwen3 gain is dominated by the \emph{comparison} category (near-chance $\sim$4\% $\to$ $\sim$42\%; Appendix~\ref{app:s3ft:per_category}) and is seed-robust ($\pm 0.4$--$1.5$\% over 3 seeds). Cross-viewpoint augmentation ($\sim$3.5$\times$ data) helps the stronger Qwen3-VL-8B base but destabilizes Qwen2.5-VL-7B; cross-base validation on Qwen2.5-VL-3B, InternVL3-8B, and LLaVA-OV-7B is in Appendix~\ref{app:s3ft:cross_base}.

\begin{table}[t]
\centering\small
\caption{SpatialBabel-QA-Score (\%) under four inference modes after Single-view S$^3$-FT (same 800 questions per cell as Table~\ref{tab:code_cot_sb}). $\Delta$ denotes the per-cell improvement over the base model.}
\label{tab:code_cot_sb_sft}
\begin{tabular}{lcccc}
\toprule
& \multicolumn{2}{c}{Qwen3-VL-8B} & \multicolumn{2}{c}{Qwen2.5-VL-7B} \\
\cmidrule(lr){2-3}\cmidrule(lr){4-5}
Mode & base & Single-view S$^3$-FT & base & Single-view S$^3$-FT \\
\midrule
direct                                    & 39.1 & \textbf{47.1}\,\deltaup{8.0} & 45.4 & \textbf{48.6}\,\deltaup{3.2} \\
nl\_cot                                   & 38.6 & \textbf{47.2}\,\deltaup{8.6} & 44.4 & \textbf{47.5}\,\deltaup{3.1} \\
best\_code\_cot (threejs)                 & 38.4 & \textbf{45.4}\,\deltaup{7.0} & 44.5 & \textbf{46.2}\,\deltaup{1.7} \\
worst\_code\_cot (canonical\_json)        & 37.5 & \textbf{42.1}\,\deltaup{4.6} & \textbf{47.1} & 38.9\,\deltadn{8.2} \\
\bottomrule
\end{tabular}
\end{table}

\nsubsection{Out-of-distribution generalization: from primitives to real photos}
\label{sec:s3ft:vlmeval}

\begin{table}[t]
\centering
\caption{General VLM performance across 12 standard benchmarks. Numbers are accuracy (\%); parenthesized values are signed deltas versus the corresponding base model (green = improvement, red = regression). \textbf{Bold} = best per row per base model. $\dagger$~MMMU re-scored with a regex letter extractor (validation split, $n{=}900$) instead of VLMEvalKit's default LLM-judge for base$\to$S$^3$-FT consistency; small deltas on this row are within re-scoring noise. S$^3$-FT is trained \emph{only} on primitive images, yet gains transfer to real-photo benchmarks.}
\label{tab:general_vl}
\small
\setlength{\tabcolsep}{3.5pt}
\begin{tabular}{lccc|ccc}
\toprule
& \multicolumn{3}{c|}{Qwen3-VL-8B} & \multicolumn{3}{c}{Qwen2.5-VL-7B} \\
Benchmark & base & Single-view & Cross-viewpoint & base & Single-view & Cross-viewpoint \\
\midrule
\multicolumn{7}{l}{\emph{Hallucination / grounding}} \\
POPE                     & \textbf{88.2} & 85.4\,\deltadn{2.8} & 86.2\,\deltadn{2.0} & \textbf{86.5} & 85.4\,\deltadn{1.1} & 85.2\,\deltadn{1.3} \\
HallusionBench (aAcc)    & 55.8 & \textbf{72.9}\,\deltaup{17.1} & 72.5\,\deltaup{16.7} & 65.2 & 68.8\,\deltaup{3.6} & \textbf{69.4}\,\deltaup{4.2} \\
\midrule
\multicolumn{7}{l}{\emph{General reasoning / multi-discipline}} \\
MMBench\textsubscript{dev-EN}     & \textbf{84.4} & 81.9\,\deltadn{2.5} & 81.3\,\deltadn{3.1} & 80.5 & \textbf{81.7}\,\deltaup{1.2} & 80.0\,\deltadn{0.5} \\
MMStar                   & \textbf{64.6} & 64.1\,\deltadn{0.5} & 63.7\,\deltadn{0.9} & 59.7 & \textbf{61.2}\,\deltaup{1.5} & 60.6\,\deltaup{0.9} \\
MMMU\textsubscript{val}$^\dagger$  & \textbf{56.8} & 56.0\,\deltadn{0.8} & 54.9\,\deltadn{1.9} & \textbf{52.4} & 51.0\,\deltadn{1.4} & 49.6\,\deltadn{2.8} \\
AI2D                     & \textbf{84.1} & 83.6\,\deltadn{0.5} & 82.6\,\deltadn{1.5} & \textbf{80.9} & 80.9\,\deltazero{0.0} & 80.6\,\deltadn{0.3} \\
\midrule
\multicolumn{7}{l}{\emph{OCR / real-world}} \\
OCRBench                 & \textbf{90.0} & 84.9\,\deltadn{5.1} & 84.4\,\deltadn{5.6} & \textbf{88.6} & 86.6\,\deltadn{2.0} & 85.3\,\deltadn{3.3} \\
RealWorldQA              & 71.0 & \textbf{72.2}\,\deltaup{1.2} & 69.0\,\deltadn{2.0} & 68.2 & \textbf{68.8}\,\deltaup{0.6} & 66.7\,\deltadn{1.5} \\
\midrule
\multicolumn{7}{l}{\emph{Spatial enrichment (out-of-distribution --- trained on primitives only)}} \\
BLINK                    & 57.1 & \textbf{59.2}\,\deltaup{2.1} & 57.5\,\deltaup{0.4} & \textbf{54.4} & 53.7\,\deltadn{0.7} & 53.4\,\deltadn{1.0} \\
CV-Bench-2D              & 70.3 & \textbf{80.0}\,\deltaup{9.7} & 78.7\,\deltaup{8.4} & 75.2 & 76.6\,\deltaup{1.4} & \textbf{77.3}\,\deltaup{2.1} \\
CV-Bench-3D              & 76.8 & \textbf{80.1}\,\deltaup{3.3} & 73.2\,\deltadn{3.6} & \textbf{71.8} & 66.6\,\deltadn{5.2} & 66.7\,\deltadn{5.1} \\
MMSIBench (non-circ.)    & \textbf{30.7} & 29.2\,\deltadn{1.5} & 25.4\,\deltadn{5.3} & \textbf{27.5} & 26.6\,\deltadn{0.9} & 25.5\,\deltadn{2.0} \\
\bottomrule
\end{tabular}
\end{table}

A critical question is whether spatial gains from primitive-only training transfer to real-photo benchmarks. Table~\ref{tab:general_vl} evaluates both base and S$^3$-FT configurations on twelve standard VLM benchmarks via VLMEvalKit~\citep{duan2024vlmevalkit} (text-only instruction-following via IFEval is in Appendix~\ref{app:s3ft:ifeval}). The results demonstrate that \textbf{primitive spatial knowledge transfers to general visual reasoning}.

For Qwen3-VL-8B under Single-view S$^3$-FT (trained \emph{only} on primitive images), most benchmarks are preserved or improved, with the spatial benchmarks showing the largest gains: HallusionBench $+17$\%, CV-Bench-2D $+9.7$\%, CV-Bench-3D $+3.3$\%, and BLINK $+2.1$\%. These are real-photo benchmarks not seen during training, and the CV-Bench and BLINK improvements constitute the primary out-of-distribution evidence that spatial knowledge learned from primitives transfers to real-photo visual reasoning. Minor regressions on POPE ($-2.8$\%) and OCRBench ($-5.1$\%) reflect the absence of OCR and hallucination supervision in the primitive-only training mixture.

For the weaker Qwen2.5-VL-7B, the same recipe yields directionally consistent but smaller gains (HallusionBench $+3$--$4$\%, CV-Bench-2D $+1.4$\%), consistent with a capacity-dependent ceiling on the spatial knowledge extractable from noisier primitive pseudo-ground-truth.

The recipe is most effective on primitive scenes, where the base model's pseudo-ground-truth is high-quality. On Hypersim photo-realistic indoor scenes, the base model's own pseudo-ground-truth covers only $\sim$38\% of objects per scene (measured on Qwen3-VL-8B and Qwen2.5-VL-7B). S$^3$-FT still lifts direct and NL-CoT modes on Hypersim ($+2.5$ and $+5.7$\% on Qwen3, $0.0$ and $+6.4$\% on Qwen2.5; Appendix Table~\ref{tab:s3ft_hypersim}), but Code-CoT modes regress ($-0.6$ to $-20.4$\% on Qwen3; $-1.9$ to $-8.3$\% on Qwen2.5), reflecting that low pseudo-ground-truth coverage on cluttered indoor scenes introduces substantial label noise on the structured-task targets that Code-CoT depends on. Improving pseudo-ground-truth quality through a stronger base model or a multi-pass aggregation strategy is a promising path forward; details are discussed in Appendix~\ref{app:s3ft:hypersim}.

%% file: sections/limitations.tex
\nsection{Limitations}
\label{sec:limitations}

S$^3$-FT is bounded by base-VLM primitive pseudo-GT quality: on Hypersim ($\sim$38\% object coverage on Qwen-VL; Appendix~\ref{app:s3ft:hypersim}) the pipeline does not transfer, and Code-CoT struggles once scene complexity exceeds what itself can capture. 
The best application domain can be therefore stylised, low-texture scenes (architectural visualizations, video-game-style graphics, robot simulators); photo-realistic transfer will likely be more challenging. 
The cross-language ranking from \textsc{SpatialBabel}-Primitive-Reconstruct (Table~\ref{tab:main_results}) also does not transfer cleanly to real photos, so the ranking is an informative default rather than a deterministic Code-CoT or S$^3$-FT language selector; we use Three.js as a single-language S$^3$-FT anchor and leave per-model language ablation to future work.

\nsection{Conclusion}
\label{sec:conclusion}

VLMs reconstruct 3D scenes from geometric primitives with high code-level fidelity yet fail at simpler spatial questions on the same images. \textsc{SpatialBabel} benchmarks this task-routing failure across six scene-code languages and fourteen VLMs, with model-specific F1 ratios up to $5.7\times$ across languages; Code-CoT routes around the failure at inference time for capable code generators, with a clear domain ceiling on cluttered photo-realistic scenes; S$^3$-FT closes it at training time for noisier coders, distilling primitive-reconstruction code into general visual reasoning with no human labeling and no teacher model. Together, 3D geometric primitives, as expressed in executable code, function as both a diagnostic instrument and a transferable spatial vocabulary for VLMs. For future work, we plan to enrich the scene-code primitive set to extend transfer into photo-realistic indoor scenes, sweep procedural complexity (occlusion, stacking, unconstrained palettes) to bridge the difficulty spectrum, and explore S$^3$-FT on stronger base VLMs. We will release the benchmark, training data, checkpoints, and evaluation toolkit upon publication. Broader impacts are discussed in Appendix~\ref{app:broader_impacts}.

%% file: sections/appendix_extras.tex
\section{Models and prompting setup}
\label{app:models_prompting}\label{sec:baselines}\label{sec:models}\label{sec:prompting}\label{sec:implementation}

We evaluate \textbf{fourteen VLMs}: eight proprietary APIs (Anthropic Claude Opus 4.7 / Sonnet 4.6, OpenAI GPT-5 / GPT-4o, Google Gemini-2.5-Pro/Flash and Gemini-3-Pro/Flash (preview)~\citep{google2025gemini}) and six open-weights bases: Qwen3-VL-8B-Instruct (the S$^3$-FT headline base), Qwen2.5-VL-7B-Instruct (second S$^3$-FT base)~\citep{bai2025qwen25vl}, Qwen2.5-VL-3B-Instruct, Qwen2.5-VL-32B-Instruct, LLaVA-OneVision-7B~\citep{li2024llavaonevision}, and InternVL3-8B (OpenGVLab). Open-weights models run on $4\times$NVIDIA A100 80GB with greedy decoding (\texttt{temperature=0.0}) for deterministic per-scene predictions; max-new-tokens 2{,}048 for code outputs (Direct reconstruction and Code-CoT), 1{,}024 for NL-CoT (which interleaves multi-sentence reasoning with a final committed answer; the lower 128-token cap used for direct answers truncated 68\% of Qwen3-VL-8B NL-CoT responses on CV-Bench-2D before answer commitment), and 256 for short direct answers. The cross-language benchmark uses two prompting modes: \textbf{Direct} (image + language-specific reconstruction instruction listing required fields) and \textbf{NL-CoT} (a natural-language reasoning prefix enumerating objects, classes, and approximate 3D positions before emitting scene code; distinct from \emph{Code-CoT} introduced later). We evaluate $n=100$ primitive scenes per cell; Hypersim uses $n=200$ per cell.

\subsection{Prompt examples}
\label{app:prompt_examples}

We list the verbatim prompt templates used throughout the paper. The reconstruction prompt is composed of a fixed system instruction, a per-language body, and dataset-specific scene-content / coordinate hints; the QA prompts (Direct, NL-CoT, Code-CoT) are constructed at evaluation time and shown below in full.

\begin{promptbox}{Reconstruction, system prompt (all languages)}
You are an expert 3D scene reconstruction assistant. Given an image of a 3D scene, you must reconstruct it as executable code in the specified programming language. Output ONLY the code, no explanations.
\end{promptbox}

\begin{promptbox}{Reconstruction, per-language body (Three.js example)}
Reconstruct this 3D scene as Three.js JavaScript code. \textit{[primitive hint]} Use the appropriate geometry (BoxGeometry, SphereGeometry, CylinderGeometry, ConeGeometry, TorusGeometry) for each object. Set exact positions, rotations, and scales. Set MeshStandardMaterial color for each object. \textit{[coordinate hint]} \textit{[camera hint]}
\end{promptbox}

The five remaining bodies (Blender Python, Unity C\#, Open3D Python, Scene Language DSL, Canonical JSON) follow the same template, swapping language-specific API verbs (e.g.\ \texttt{primitive\_*\_add()}, \texttt{GameObject.CreatePrimitive()}, \texttt{create\_*}, \texttt{(entity ...)}). The Canonical JSON body inlines an explicit schema:

\begin{promptbox}{Reconstruction, per-language body (Canonical JSON)}
Reconstruct this 3D scene as a JSON object with this exact schema: \{"scene\_id": "predicted", "objects": [\{"class\_name": "shape", "position": [x,y,z], "rotation": [rx,ry,rz], "scale": [sx,sy,sz], "material": "color\_name"\}], "camera": \{"position": [x,y,z], "target": [x,y,z], "fov": 60.0\}\}
\end{promptbox}

\paragraph{Reconstruction, dataset hints (appended to body).}
\textbf{Primitive hint:} ``\texttt{Objects are geometric primitives: cube, sphere, cylinder, cone, torus, pyramid. Each has a color from: red, blue, green, yellow, purple, orange, cyan, white.}''
\textbf{Hypersim hint:} ``\texttt{This is a photorealistic indoor scene. Objects are real furniture and items in a room. Use descriptive class names for each object (e.g., chair, table, bathtub, lamp, bed).}''
\textbf{Coordinate hint:} a Y-up right-handed convention plus per-axis spatial extent (e.g.\ ``\texttt{The ground plane extends from $-3.0$ to $+3.0$ in both X and Z.}'' for primitives; per-scene bounds for Hypersim) and a ground-plane convention (\texttt{y=0} with each object's $y$ position equal to half its height for primitives; floor at $Y\!\approx\!0$ in meters for Hypersim).

\begin{promptbox}{Spatial QA, Direct mode (Tables 2 to 4, \texttt{direct} column)}
\{question\}\\
Answer concisely.
\end{promptbox}

\begin{promptbox}{Spatial QA, NL-CoT mode (Tables 2 to 4, \texttt{nl\_cot} column)}
\{question\}\\
\\
Think step-by-step about the objects in the scene, their classes, colours, and approximate 3-D positions. Then give a concise final answer.
\end{promptbox}

\begin{promptbox}{Spatial QA, Code-CoT mode (Tables 2 to 4, \texttt{best\_cc} / \texttt{worst\_cc} columns)}
First, write a complete \{language\} reconstruction of the 3-D scene you see. Include each object's class, colour, and position. Then, using your reconstruction, answer the following question.\\
\\
Question: \{question\}\\
Finish with a line starting with `Final answer:' followed by the concise answer.
\end{promptbox}

\texttt{\{language\}} is replaced with a human-readable rendering of each model's primitive-best or primitive-worst scene-code language (e.g.\ ``\texttt{Three.js JavaScript}'', ``\texttt{canonical JSON with the schema \{...\}}'').

\section{Scene-code language exposure proxies}
\label{app:exposure_proxies}

A key contribution of \textsc{SpatialBabel} is evaluating spatial understanding across six diverse 3D scene-code languages, spanning different paradigms and syntax families: \textbf{Three.js} (imperative web 3D), \textbf{Canonical JSON} (declarative structured), \textbf{Blender Python} (procedural scripting), \textbf{Unity C\#} (engine API with transform components), \textbf{Open3D Python} (mesh-creation API), and a \textbf{Scene Language DSL}~\cite{zhang2025scene}. They differ substantially in public footprint:

\begin{table}[h]
\centering
\caption{Public proxy metrics for pre-training exposure (collected 2026-04). Languages ordered by approximate pre-training prevalence (most $\rightarrow$ least). Canonical JSON is introduced by this work, so specific-schema exposure is zero, though generic JSON priors remain available to all models.}
\label{tab:exposure_proxies}
\small
\setlength{\tabcolsep}{4pt}
\begin{tabular}{lrrrr}
\toprule
Language & First release & Pkg downloads & SO questions & GH stars \\
\midrule
Three.js         & 2010 & 8.08M/wk (npm) & 21{,}057 & 112.0k \\
Unity C\#        & 2005 & closed-source  & 77{,}706 & 12.8k \\
Blender Python   & 2011 & 16k/wk (PyPI)  & 3{,}313  & 18.1k \\
Open3D Python    & 2018 & 372k/wk (PyPI) & 422      & 13.5k \\
Canonical JSON   & 2026 (ours) & ---      & ---      & ---    \\
Scene Lang.\ DSL & 2024 & ---            & 0        & 257    \\
\bottomrule
\end{tabular}
\end{table}

We treat this exposure axis as a continuous variable for analysis (Section~\ref{sec:analysis}) rather than forcing a binary grouping, because model-specific behavior does not cleanly decompose into two tiers.

\section{Code-CoT setup details and QA evaluator}
\label{app:cot_evaluator}

This appendix expands on the Code-CoT setup paragraph (Section~\ref{sec:code_cot:setup}).

\textbf{Direct mode for proprietary APIs is not literally ``no reasoning''.} For frontier API models (Claude Opus 4.7, Claude Sonnet 4.6, GPT-5, GPT-4o, and the four Gemini variants), direct mode still executes internal reasoning tokens before the visible response; NL-CoT and Code-CoT therefore mainly change the \emph{visible} reasoning scaffold for these models. For open-weights bases (Qwen3-VL-8B, Qwen2.5-VL-7B/3B, InternVL3-8B, LLaVA-OV-7B), sampled direct-mode outputs average 5 to 7 characters with no \texttt{<think>} traces, so chain-of-thought is genuinely absent under direct.

\textbf{NL-CoT prompt.} Our NL-CoT prompt asks the model to think step-by-step about object class, color, and approximate 3D position before producing a final answer; this is an information-matched prose analogue of canonical-JSON Code-CoT, deliberately strong so that any Code-CoT advantage cannot be attributed to merely cuing the model to attend to object attributes.

\textbf{Localization category.} The localization category in SpatialBabel-QA requires emitting numerical $(x,y,z)$ coordinates within tight tolerance, which is intractable for current numerical-extraction parsers and consistently scores near zero across all models and modes; the overall accuracies reported in Section~\ref{sec:code_cot} are roughly uniformly dragged down by this.

\textbf{CoT-symmetric QA evaluator.} The QA scorer (\texttt{evaluate\_qa\_answer}) applies the same five category parsers uniformly across direct, NL-CoT, and Code-CoT, eliminating prompt-format asymmetries that earlier paper drafts inherited from full-response substring matching.

\textbf{Per-category extraction.} (i)~\emph{Counting:} prefer the integer following an explicit ``Final answer:'' / ``the answer is:'' / ``total:'' marker; otherwise take the last integer in the response (chain-of-thought scratchpads typically commit at the end). (ii)~\emph{Existence:} prefer the \emph{yes}/\emph{no} token following an explicit answer marker; otherwise take the last \emph{yes}/\emph{no} word-boundary token. (iii)~\emph{Relationship} and (iv)~\emph{comparison:} extract the trailing committed answer segment (the final ``Final answer:'' / ``Answer:'' line, or the last 1 or 2 non-empty lines) and substring-match the ground truth phrase against \emph{that} segment, not the full response. (v)~\emph{Localization:} parse the first three numbers in the response (unchanged from earlier drafts).

\textbf{Code-CoT pre-extraction.} For Code-CoT modes, we additionally apply a markdown-tolerant ``Final answer:'' line extractor before passing to \texttt{evaluate\_qa\_answer}, so both Code-CoT and NL-CoT scoring see the model's committed answer rather than its scratchpad.

The CoT-symmetric variant brings NL-CoT scoring strictness in line with Code-CoT scoring strictness so the inference-mode comparison in Tables~\ref{tab:code_cot_sb}, \ref{tab:code_cot_hypersim}, and~\ref{tab:code_cot_cvbench} is apples-to-apples.

\section{Code-CoT on SpatialBabel-Hypersim-QA}
\label{app:code_cot_hypersim}

Table~\ref{tab:code_cot_hypersim} reports the four-mode SpatialBabel-Hypersim-QA evaluation (157 questions per cell, 20 held-out photo-realistic indoor scenes). Primitive-best Code-CoT is sign-mixed: five of ten models gain (Gemini-2.5-Flash $+6.4$\%, Gemini-2.5-Pro $+5.1$\%, Sonnet $+4.5$\%, Gemini-3-Flash $+3.8$\%, Qwen2.5-VL-7B $+2.5$\%) and five regress (Opus 4.7 $-5.1$\%, GPT-4o $-3.8$\%, Gemini-3-Pro $-3.8$\%, GPT-5 $-3.2$\%, Qwen3-VL-8B $-1.9$\%). Under the CoT-symmetric evaluator the row maxima split evenly across direct (2 of 10), NL-CoT (4 of 10), and Code-CoT (4 of 10). The primitive-best language is also a leaky predictor on cluttered indoor scenes (e.g., GPT-4o's primitive-worst DSL beats its primitive-best Three.js by $8.9$pp), an inversion that recurs on CV-Bench-3D and is discussed in Section~\ref{sec:limitations}.

\begin{table}[h]
\centering\small
\caption{SpatialBabel-Hypersim-QA-Score (\%) under four inference modes. \texttt{best\_cc} / \texttt{worst\_cc} use each model's primitive-best / primitive-worst scene-code language from Table~\ref{tab:main_results}. $\Delta_\text{best-direct} = \texttt{best\_cc} - \texttt{direct}$ (pp). Bold = row max.}
\label{tab:code_cot_hypersim}
\begin{tabular}{lccccc}
\toprule
Model              & direct & nl\_cot & best\_cc & worst\_cc & $\Delta_\text{best-direct}$ \\
\midrule
\multicolumn{6}{l}{\emph{Proprietary APIs}} \\
Claude Opus 4.7    & \textbf{30.6} & 28.7          & 25.5          & 26.8          & \deltadn{5.1} \\
Claude Sonnet 4.6  & 21.7          & 29.3          & 26.1          & \textbf{29.9} & \deltaup{4.5} \\
GPT-5              & 25.5          & \textbf{26.1} & 22.3          & 23.6          & \deltadn{3.2} \\
GPT-4o             & 24.8          & 26.8          & 21.0          & \textbf{29.9} & \deltadn{3.8} \\
Gemini-2.5-Pro     & 24.2          & \textbf{29.9} & 29.3          & 29.3          & \deltaup{5.1} \\
Gemini-2.5-Flash   & 24.8          & 29.3          & \textbf{31.2} & 28.7          & \deltaup{6.4} \\
Gemini-3-Flash     & 24.8          & \textbf{33.1} & 28.7          & 26.8          & \deltaup{3.8} \\
Gemini-3-Pro       & 29.3          & \textbf{30.6} & 25.5          & 29.3          & \deltadn{3.8} \\
\midrule
\multicolumn{6}{l}{\emph{Open weights}} \\
Qwen3-VL-8B        & \textbf{25.5} & 21.7          & 23.6          & 14.0          & \deltadn{1.9} \\
Qwen2.5-VL-7B      & 28.0          & 22.9          & \textbf{30.6} & 21.7          & \deltaup{2.5} \\
\bottomrule
\end{tabular}
\end{table}

\section{CV-Bench Code-CoT, 2D and 3D (out-of-distribution)}
\label{app:cvbench_table}\label{sec:code_cot:cvbench}

SpatialBabel's questions are programmatically generated, so they favour code-CoT by construction. To test whether the lift generalises to real-photo spatial reasoning, we repeat the four-mode comparison on CV-Bench-2D and CV-Bench-3D~\citep{cvbench2024} (1{,}438 and 1{,}200 real-photo MCQs respectively); the per-cell results are in Table~\ref{tab:code_cot_cvbench}. Two patterns separate the two splits. \textbf{(i) Code-CoT lifts on 3D where it does not on 2D.} On CV-Bench-2D, primitive-best Code-CoT changes accuracy by less than $\pm 4$\% relative to direct for every model (Claude Sonnet 4.6 $+2.2$\% and Gemini-3-Flash $+0.4$\% are the only positive $\Delta$; the rest range from $0$ to $-3.8$\%). On CV-Bench-3D the picture is more positive: best\_cc is the row maximum for Claude Opus 4.7 ($89.0$\%) and a strong second on most other rows, and nl\_cot helps 7 of 8 frontier models on 3D (gains up to $+7.4$\% for Claude Sonnet 4.6; only Gemini-2.5-Pro slightly regresses by $-0.4$\%). Depth/distance reasoning benefits from explicit reasoning trace in a way that 2D primitives do not. \textbf{(ii) Primitive-best is again leaky on real photos.} For Gemini-3-Pro ($96.0$\% via Scene Lang DSL vs.\ $95.1$\% via Three.js on CV-Bench-3D), Gemini-3-Flash ($94.3$\% via Unity C\# vs.\ $93.6$\% via Canonical JSON on CV-Bench-3D), GPT-4o ($83.0$\% DSL vs.\ $78.2$\% Three.js on CV-Bench-3D), and Claude Sonnet 4.6 ($87.3$\% DSL vs.\ $86.3$\% JSON on CV-Bench-3D), the primitive-\emph{worst} Code-CoT language exceeds the primitive-best on real photos, the same inversion observed on Hypersim (Section~\ref{sec:code_cot:hypersim}). The primitive Reconstruct-Score ranking is a useful default but not a tight predictor of Code-CoT effectiveness on cluttered real-photo benchmarks.

\begin{table}[h]
\centering\small
\caption{Accuracy (\%) under four inference modes on CV-Bench-2D (1{,}438 items) and CV-Bench-3D (1{,}200 items). \texttt{best\_cc} and \texttt{worst\_cc} use each model's primitive-best and primitive-worst scene-code language from Table~\ref{tab:main_results}, matching the convention of Tables~\ref{tab:code_cot_sb} and~\ref{tab:code_cot_hypersim}. \textbf{Bold} = row maximum within each benchmark. The Qwen3-VL-8B CV-Bench-2D \texttt{best\_cc} cell (Canonical JSON, its primitive-best) is left blank because the full 1{,}438-item run was not completed in time; the corresponding CV-Bench-3D cell (1{,}200 items) was run and is reported.}
\label{tab:code_cot_cvbench}
\resizebox{\linewidth}{!}{%
\begin{tabular}{lcccc|cccc}
\toprule
& \multicolumn{4}{c|}{CV-Bench-2D (1{,}438)} & \multicolumn{4}{c}{CV-Bench-3D (1{,}200)} \\
\cmidrule(lr){2-5}\cmidrule(lr){6-9}
Model & direct & nl\_cot & best\_cc & worst\_cc & direct & nl\_cot & best\_cc & worst\_cc \\
\midrule
\multicolumn{9}{l}{\emph{Proprietary APIs}} \\
Claude Opus 4.7    & 80.4 & 80.6 & 80.3 & \textbf{80.9} & 84.0 & 87.8 & \textbf{89.0} & 88.1 \\
Claude Sonnet 4.6  & 76.8 & 78.4 & \textbf{79.0} & 78.3          & 80.4 & \textbf{87.8} & 86.3 & 87.3 \\
GPT-5              & \textbf{80.9} & 80.8 & 80.2 & 77.2          & 83.8 & \textbf{89.6} & 87.0 & 80.4 \\
GPT-4o             & 74.4 & 78.4 & 70.6 & \textbf{80.0}          & 78.5 & \textbf{83.7} & 78.2 & 83.0 \\
Gemini-2.5-Pro     & 80.9 & \textbf{81.7} & 80.0 & 79.6          & \textbf{92.2} & 91.8 & 91.3 & 89.9 \\
Gemini-2.5-Flash   & \textbf{81.7} & 81.4 & 80.8 & 80.5          & 88.9 & \textbf{90.9} & 87.1 & 87.5 \\
Gemini-3-Flash     & 84.6 & 83.9 & 85.0 & \textbf{85.3}          & 92.8 & 93.6 & 93.6 & \textbf{94.3} \\
Gemini-3-Pro       & 84.1 & 84.8 & 84.1 & \textbf{85.5}          & 95.5 & 95.8 & 95.1 & \textbf{96.0} \\
\midrule
\multicolumn{9}{l}{\emph{Open weights}} \\
Qwen3-VL-8B        & \textbf{80.4} & 79.7 & ---  & 70.0          & \textbf{92.9} & 92.5 & 83.2 & 65.1 \\
Qwen2.5-VL-7B      & \textbf{75.9} & 53.3 & 73.4 & 75.5          & \textbf{82.7} & 76.9 & 71.2 & 77.5 \\
\bottomrule
\end{tabular}}
\end{table}

\section{Per-scene-code Reconstruct Score on Hypersim}
\label{app:hypersim_table}

\begin{table}[h]
\centering
\caption{Per-scene-code SpatialBabel-Reconstruct-Score on \textbf{Hypersim} indoor scenes (FOV-filtered, $n=200$ scenes per row) for the five models with full Hypersim coverage. Format mirrors Table~\ref{tab:main_results} for direct primitive$\leftrightarrow$Hypersim comparison: \textbf{bold} = best per model-row, \colorbox{red!15}{red} = lowest per model-row, $\Delta$ = max minus min Reconstruct Score gap. Hypersim numbers are FOV-filtered (GT objects whose 3D centre falls outside the rendered camera frustum are excluded; Section~\ref{sec:limitations}); mean GT object count is 65.7/scene unfiltered, 61.3/scene FOV-filtered. The Reconstruct Score is dragged down on Hypersim primarily by class accuracy (60-class indoor taxonomy vs.\ primitive's 6 shapes) and scale fidelity (indoor objects span a wider size range).}
\label{tab:hypersim}
\small
\setlength{\tabcolsep}{4pt}
\begin{tabular}{l cccccc c}
\toprule
Model & Three.js & Unity & Blen.Py & Open3D & JSON & DSL & $\Delta$ \\
\midrule
\multicolumn{8}{l}{\emph{Proprietary APIs}} \\
Claude Opus 4.7 & .444 & \colorbox{red!15}{.433} & .452 & .444 & \textbf{.458} & .453 & .025 \\
GPT-5           & .439 & \colorbox{red!15}{.350} & .370 & .414 & \textbf{.450} & .403 & .100 \\
Gemini-2.5-Pro  & .372 & \colorbox{red!15}{.329} & .351 & .381 & \textbf{.405} & .347 & .077 \\
\midrule
\multicolumn{8}{l}{\emph{Open weights}} \\
Qwen2.5-VL-7B   & .403 & .362 & \textbf{.429} & .347 & .312 & \colorbox{red!15}{.279} & .149 \\
Qwen3-VL-8B     & .414 & .425 & .429 & .442 & \textbf{.456} & \colorbox{red!15}{.401} & .055 \\
\bottomrule
\end{tabular}
\\\smallskip
{\footnotesize \textbf{Primitive $\to$ Hypersim drop} (mean of per-language Direct SpatialBabel-Reconstruct-Score): Claude Opus 4.7 $0.710 \to 0.447$ ($-0.263$); GPT-5 $0.608 \to 0.404$ ($-0.204$); Gemini-2.5-Pro $0.637 \to 0.364$ ($-0.273$); Qwen2.5-VL-7B $0.463 \to 0.355$ ($-0.108$); Qwen3-VL-8B $0.621 \to 0.428$ ($-0.193$). Per-row best/worst languages are largely preserved across splits (Canonical JSON is each proprietary model's Hypersim-best, matching Table~\ref{tab:main_results}'s pattern for the Sonnet/GPT-5/Gemini-Flash family on primitives, and Unity C\# is the proprietary-row Hypersim-worst), supporting the claim in Section~\ref{sec:analysis} that the cross-scene-code gap tracks language familiarity rather than scene content.}
\end{table}

\section{SpatialBabel-Reconstruct-Score per-component breakdown}
\label{app:sbs_components}

Table~\ref{tab:sbs_components} reports the five Reconstruct Score components averaged across the six scene-code languages for each (model, mode) cell. The Reconstruct Score is their arithmetic mean (last column).

\begin{table}[h]
\centering\small
\setlength{\tabcolsep}{4.5pt}
\caption{SpatialBabel-Reconstruct-Score components, per (model, mode), averaged across the six scene-code languages on primitive scenes. \textbf{Parse} = parse rate; \textbf{F1} = Hungarian-matched object F1; \textbf{ClsAcc} = class accuracy on matched pairs; \textbf{PosFid} = $1{-}\,\text{clipped pos\_rmse\_norm}$; \textbf{ScaleFid} = $1{-}\,\text{clipped scale\_rmse}$.}
\label{tab:sbs_components}
\begin{tabular}{llcccccc}
\toprule
Model & Mode & Parse & F1 & ClsAcc & PosFid & ScaleFid & Mean \\
\midrule
\multirow{2}{*}{Gemini-2.5-Pro} & base   & 0.814 & 0.761 & 0.382 & 0.572 & 0.654 & \textbf{0.636} \\
                                & NL-CoT & 0.931 & 0.864 & 0.419 & 0.594 & 0.663 & \textbf{0.694} \\
\midrule
\multirow{2}{*}{Qwen2.5-VL-32B} & base   & 0.856 & 0.747 & 0.311 & 0.501 & 0.612 & \textbf{0.605} \\
                                & NL-CoT & 0.922 & 0.775 & 0.296 & 0.497 & 0.594 & \textbf{0.617} \\
\midrule
\multirow{2}{*}{Qwen3-VL-8B}    & base   & 0.908 & 0.726 & 0.273 & 0.493 & 0.705 & \textbf{0.621} \\
                                & NL-CoT & 0.797 & 0.666 & 0.275 & 0.476 & 0.735 & \textbf{0.590} \\
\midrule
\multirow{2}{*}{Qwen2.5-VL-7B}  & base   & 0.597 & 0.439 & 0.186 & 0.384 & 0.706 & \textbf{0.463} \\
                                & NL-CoT & 0.802 & 0.665 & 0.226 & 0.409 & 0.709 & \textbf{0.562} \\
\midrule
\multirow{2}{*}{LLaVA-OV-7B}    & base   & 0.851 & 0.694 & 0.265 & 0.445 & 0.766 & \textbf{0.604} \\
                                & NL-CoT & 0.761 & 0.596 & 0.203 & 0.455 & 0.738 & \textbf{0.550} \\
\midrule
\multirow{2}{*}{Qwen2.5-VL-3B}  & base   & 0.657 & 0.531 & 0.199 & 0.440 & 0.723 & \textbf{0.510} \\
                                & NL-CoT & 0.842 & 0.630 & 0.202 & 0.426 & 0.639 & \textbf{0.548} \\
\bottomrule
\end{tabular}
\end{table}

\paragraph{What the breakdown reveals.} The table makes the failure modes explicit. (i) \emph{Parse rate} (fraction of outputs that yield $\geq 1$ object after parsing) ranges from $\sim 0.60$ on the weakest 7B/3B Qwen bases up to $\sim 0.91$ on Qwen3-VL-8B; Gemini-2.5-Pro sits at $0.81$ owing to a small fraction of refusals on JSON output. The dominant failure on the weak Qwen bases is empty parses (valid syntax, zero objects extracted), not malformed code. (ii) \emph{Class accuracy} on matched pairs is the largest cross-model gap: Gemini reaches $0.38$ averaged across languages while open-weights bases sit around $0.19$ to $0.31$. (iii) \emph{Position fidelity} cleanly separates models that learn the GT coordinate frame (Gemini 0.57) from those that emit camera-relative outputs (Qwen3-VL-8B 0.49). (iv) \emph{Scale fidelity} is high across the board on primitive scenes because most primitives default to scale $\sim 1$, so the metric is loose; it tightens substantially on Hypersim where object sizes span a much wider range.

\section{Reconstruct Score vs QA Score correlation}
\label{app:sbs_qa_corr}

The headline finding in Section~\ref{sec:finding_reconstruct_vs_qa} is that primitive-reconstruction skill (Reconstruct-Score F1) and direct spatial answering (QA Score) measure complementary, weakly-correlated capabilities. Table~\ref{tab:sbs_qa_corr} reports the per-model values used to compute the correlations.

\begin{table}[h]
\centering\small
\setlength{\tabcolsep}{6pt}
\caption{Per-model comparison of Reconstruct-Score F1 (Hungarian-matched object F1, averaged across the six scene-code languages, Direct prompting) against SpatialBabel-Primitive-QA accuracy on the relationship category and overall (100 scenes $\times$ 8 questions, Direct prompting). Across the nine VLMs evaluated on both, Reconstruct F1 spans $0.439$ to $0.937$ (a $2.13\times$ ratio) while QA-relationship clusters within a $2.5$-pp band. Pearson $r$ and Spearman $\rho$ at the bottom; per-scene-code correlations of Reconstruct F1 against QA-relationship report which scene-code languages' reconstruction skill is most predictive of relational reasoning.}
\label{tab:sbs_qa_corr}
\begin{tabular}{lccc}
\toprule
Model & Reconstruct F1 (mean of 6 langs) & QA-relationship (\%) & QA-overall (\%) \\
\midrule
Claude Opus 4.7   & \textbf{.937} & \textbf{38.2} & \textbf{48.8} \\
Gemini-3-Pro      & .904          & 37.9 & 47.2 \\
GPT-4o            & .888          & 37.3 & 43.0 \\
Claude Sonnet 4.6 & .867          & 37.9 & 43.1 \\
Gemini-3-Flash    & .807          & 37.3 & 47.4 \\
Gemini-2.5-Pro    & .761          & \textbf{38.2} & 45.5 \\
Gemini-2.5-Flash  & .726          & 37.6 & 48.1 \\
Qwen3-VL-8B       & .726          & \colorbox{red!15}{35.7} & \colorbox{red!15}{39.1} \\
Qwen2.5-VL-7B     & \colorbox{red!15}{.439} & 37.9 & 45.4 \\
\midrule
\multicolumn{2}{l}{\emph{Reconstruct F1 (mean) vs.\ QA-relationship}} & Pearson $r=0.12$ & Spearman $\rho=0.10$ \\
\multicolumn{2}{l}{\emph{Reconstruct F1 (mean) vs.\ QA-overall}} & Pearson $r=0.15$ & Spearman $\rho=0.23$ \\
\midrule
\multicolumn{4}{l}{\emph{Per-scene-code Reconstruct F1 vs.\ QA-relationship} (Pearson $r$ / Spearman $\rho$, $n=9$):} \\
\multicolumn{4}{l}{\hspace{1em}Open3D Py $+0.50 / +0.40$ \,$\cdot$\, Blender Py $+0.33 / +0.17$ \,$\cdot$\, Three.js $+0.22 / +0.18$} \\
\multicolumn{4}{l}{\hspace{1em}Unity C\# $-0.15 / -0.10$ \,$\cdot$\, Canon. JSON $-0.05 / -0.30$ \,$\cdot$\, Scene DSL $-0.24 / -0.30$} \\
\bottomrule
\end{tabular}
\end{table}

Two patterns separate this table from a typical benchmark-correlation analysis. \textbf{(i) Reconstruction range $\gg$ relational-QA range.} Reconstruct F1 spans more than $2\times$ across the nine models while QA-relationship spans $2.5$ percentage points; the ranges differ by an order of magnitude. \textbf{(ii) Geometric-API scene-code languages are the only modest predictors.} Open3D Python ($+0.50$) and Blender Python ($+0.33$) show positive correlation with relational-QA, while Scene Language DSL is mildly \emph{negative} ($-0.24$), consistent with DSL being a flat declarative format that does not require explicit spatial reasoning during code generation. The two patterns together suggest the relational-QA bottleneck is upstream of code-generation skill: the model that reconstructs primitive scenes more faithfully has more spatial \emph{evidence}, but does not by that fact alone reach the relational \emph{conclusion} when asked directly.

\section{Additional benchmark visualizations and findings}
\label{app:benchmark_extras}

\subsection{Per-scene-code Reconstruct Score bar charts}
\label{app:per_language_bars}

\begin{figure}[h]
\centering
\includegraphics[width=0.85\linewidth]{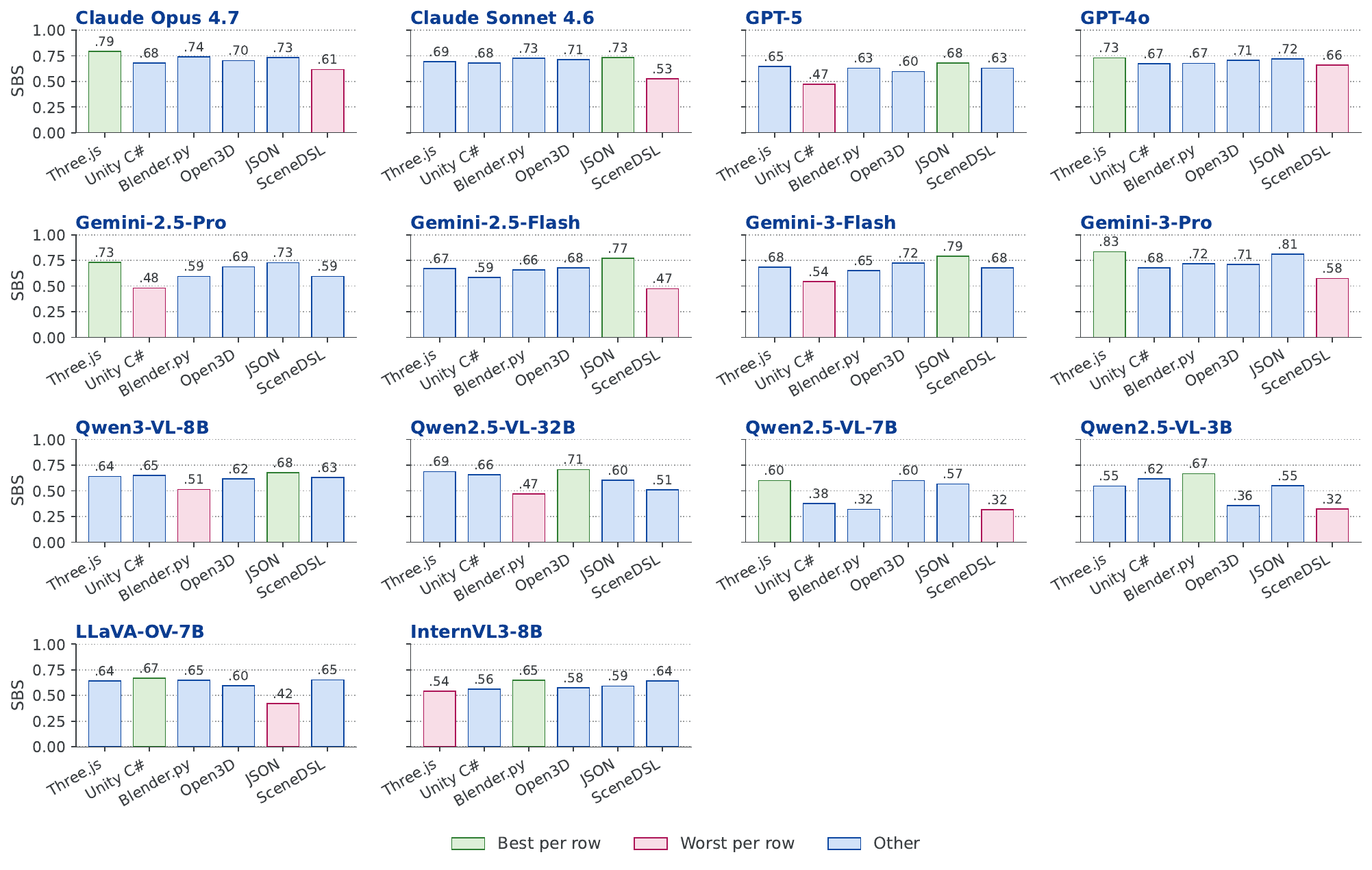}
\caption{Per-scene-code SpatialBabel-Reconstruct-Score on primitive scenes under \emph{Direct} prompting per model. Proprietary APIs are in rows 1 and 2; open-weights bases in rows 3 and 4. \textcolor[HTML]{2E7D32}{Green} bars = best language per row, \textcolor[HTML]{AD1457}{red} bars = worst per row, blue bars = other (matching Table~\ref{tab:main_results}'s bold/red convention). Same data as Table~\ref{tab:main_results}.}
\label{fig:per_language_f1}
\end{figure}

\subsection{Language-difficulty heatmap and scale findings}
\label{app:scale_findings}

\begin{figure}[h]
\centering
\includegraphics[width=0.95\linewidth]{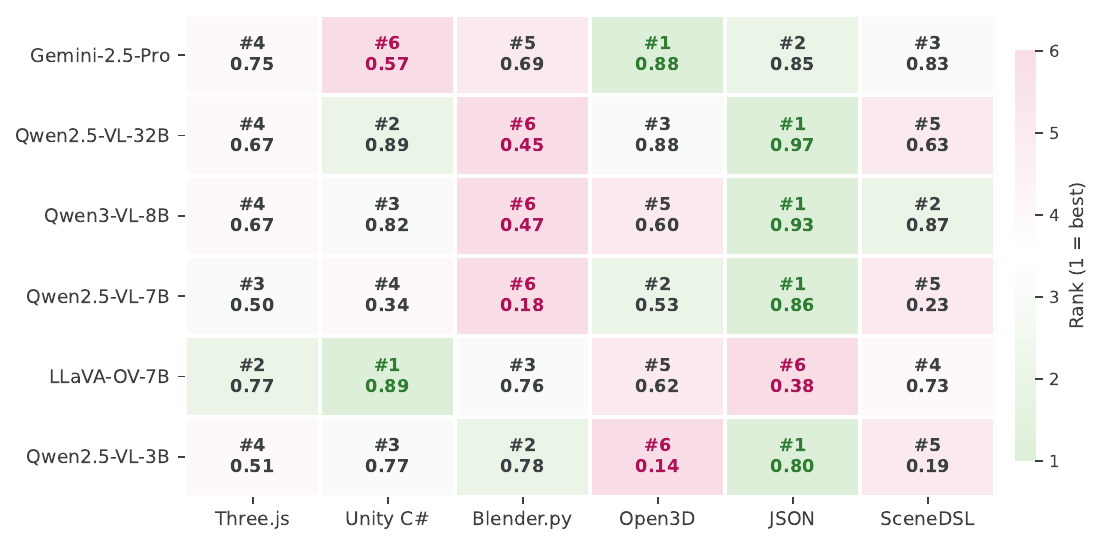}
\caption{Language difficulty rankings per model (baseline, primitive scenes). Each cell shows the model's rank among the 6 languages (1 = best) and absolute F1. Models disagree on which languages are easy or hard, and the ranking pattern correlates with model scale: smaller models (Qwen2.5-VL-3B, LLaVA-OV-7B) rank high-exposure languages above low-exposure ones, whereas larger models show the opposite pattern driven primarily by Canonical JSON being structurally simple.}
\label{fig:language_rankings}
\end{figure}

\paragraph{The benchmark scales with model size.} Within the Qwen2.5 family (a controlled three-point sweep at 3B, 7B, and 32B parameters), mean F1 across all six languages rises monotonically with parameter count. Under Direct: 0.463 (3B), 0.439 (7B), 0.744 (32B); under NL-CoT: 0.633 / 0.665 / 0.775. The 3B$\to$32B step ($10\times$ parameter count) lifts mean F1 by $\sim$0.28 (Direct) and $\sim$0.14 (NL-CoT). The monotonic-with-scale ordering is a sanity check for the benchmark itself: a benchmark that did not separate model sizes would suggest its signal was dominated by something other than spatial-understanding capacity.

\paragraph{Scale changes \emph{where} the gaps appear, not just their magnitude.} Smaller models (Qwen2.5-VL-3B, LLaVA-OV-7B) underperform on less-exposed languages such as Open3D and Scene Language DSL, the intuitive failure mode. Larger models (Qwen2.5-VL-7B with NL-CoT, Gemini-2.5-Pro) instead tend to fail on specific widely-exposed APIs while excelling on novel formats: Qwen2.5-VL-7B scores 0.245 on Blender Python under NL-CoT while reaching 0.875 on Scene Language DSL; Gemini scores 0.810 on Three.js (its lowest F1) while reaching 0.995 on Canonical JSON. Small models are exposure-bound; large models have absorbed all six APIs but still show idiosyncratic weak spots driven by API \emph{semantics} rather than aggregate public code frequency. In neither regime does scale eliminate cross-language variance.

\subsection{Same-scene cross-language qualitative comparison}
\label{app:cross_lang_qualitative}

\begin{figure}[h]
\centering
\includegraphics[width=\linewidth]{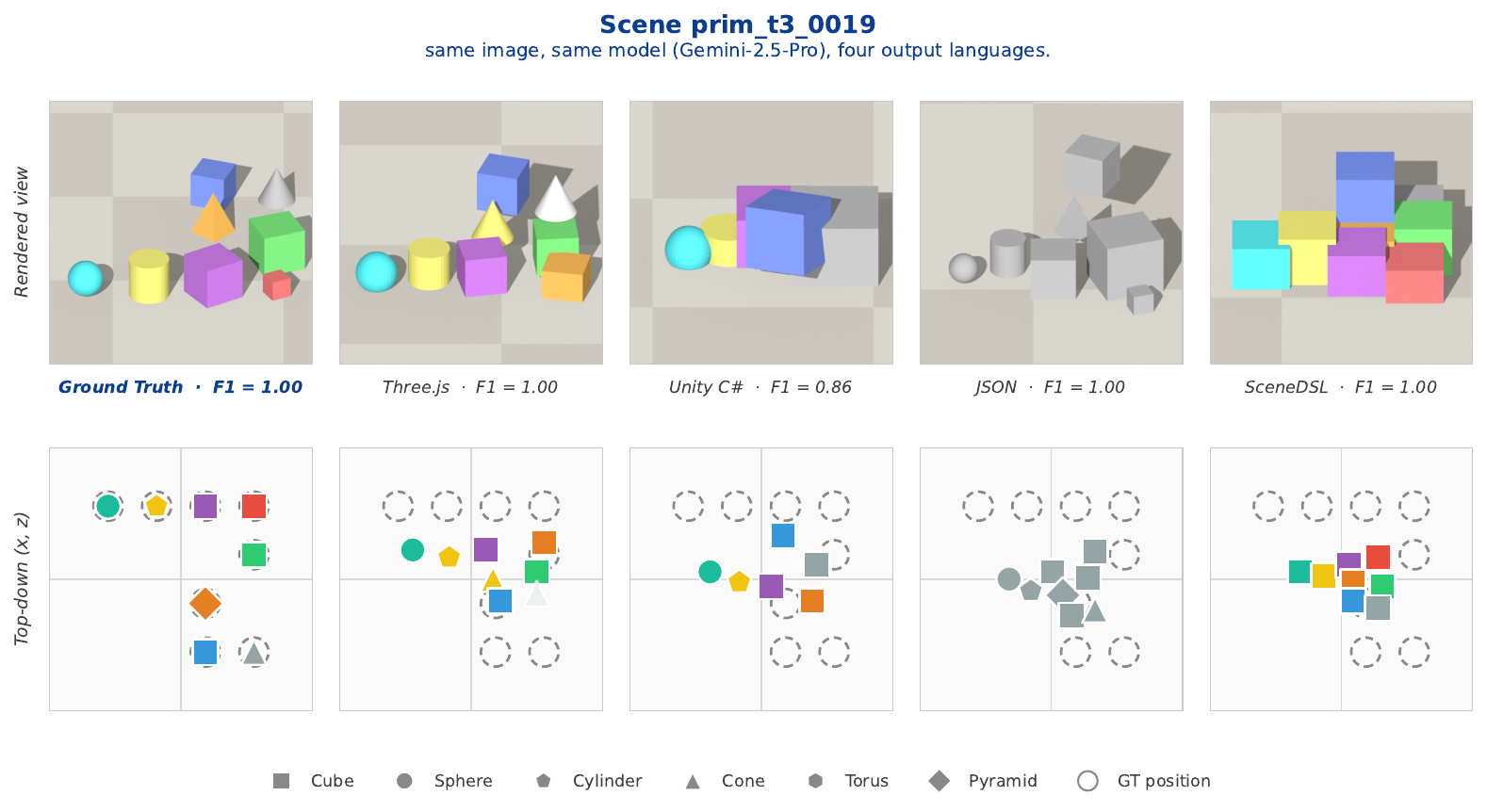}
\caption{Same scene, same model (Gemini-2.5-Pro), four scene-code languages. \textbf{Top row:} 3D render of the ground truth and of each predicted scene, reconstructed from the model's generated code and rendered with the GT camera. \textbf{Bottom row:} top-down (x,z) layout; dashed rings = GT positions, filled markers = predicted positions. Gemini's Unity C\# output (F1=0.86) recovers the scene with slightly different object placement than its Three.js output (F1=1.00) despite both describing the same image.}
\label{fig:qualitative_gemini_langs}
\end{figure}

\begin{figure}[h]
\centering
\includegraphics[width=0.95\linewidth]{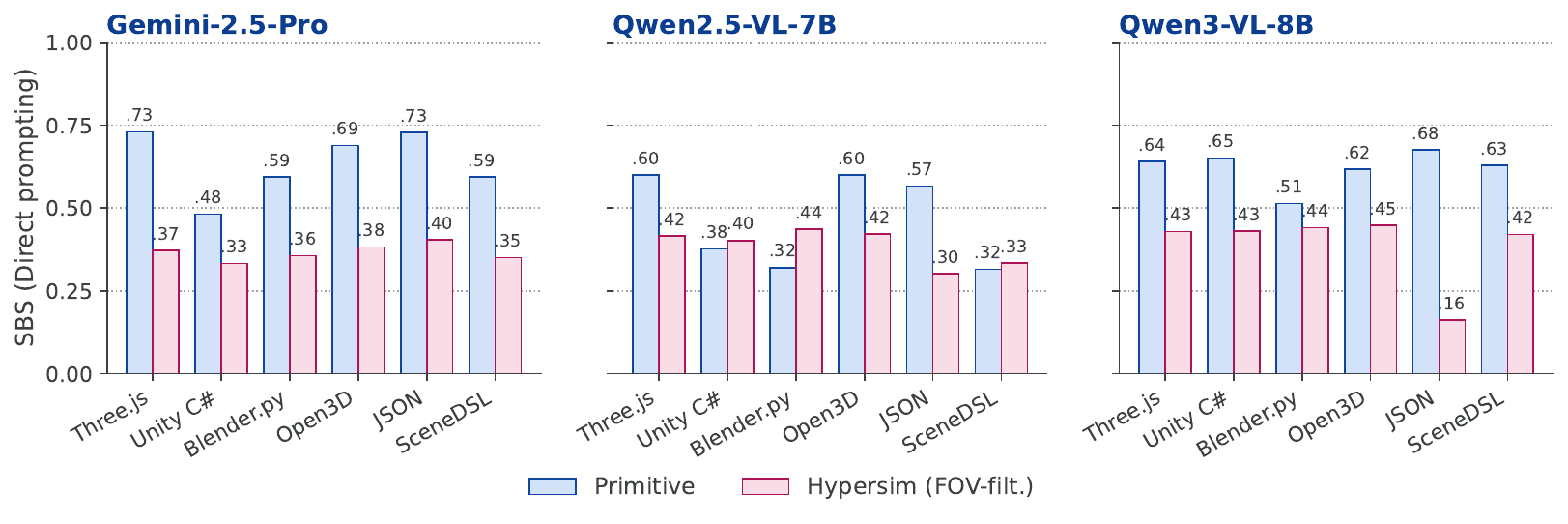}
\caption{Primitive vs.\ Hypersim per-scene-code Reconstruct Score (best of Direct/NL-CoT) for the three models with both splits. Hypersim numbers are FOV-filtered (full per-component metrics computed against FOV-matched pairs), matching Table~\ref{tab:hypersim}. Hypersim scenes are substantially harder across all languages, but the per-language profile is preserved: cross-language gaps are driven by language familiarity rather than scene content.}
\label{fig:primitive_vs_hypersim}
\end{figure}

\section{S\texorpdfstring{$^3$}{3}-FT additional analyses}
\label{app:s3ft_extras}

This appendix contains the S$^3$-FT analyses not on the main body's critical-path narrative. The main paper covers the pipeline (Section~\ref{sec:s3ft:pipeline}), the Single-view recipe with in-distribution gains (Section~\ref{sec:s3ft:single-view}), and the out-of-distribution VLMEvalKit performance comparison (Section~\ref{sec:s3ft:vlmeval}). Below we report: the v1 format-collapse failure mode that motivates Single-view, the Cross-viewpoint augmentation variant, full per-language and per-category breakdowns, multi-seed reliability, an ablation of the design choices, the Hypersim-domain failure case, and cross-base/cross-family validation.

\subsection{Format-collapse failure mode (v1)}
\label{app:s3ft:collapse}

A naive instantiation of the S$^3$-FT pipeline (we call it \textbf{v1}) trains only on the $K=9$ structured tasks derived in Phase 2 with no free-form supervision. v1 dramatically improves structured outputs but \emph{collapses} the model's language-level behaviour: on 100 held-out primitive scenes, mean benchmark F1 drops from $0.428$ to $0.280$ for Qwen2.5-VL-7B; SpatialBabel-QA drops $-3.3$\% to $38.6\%$. Inspecting failures, we find the fine-tuned model emits scene-graph JSON for any prompt, even free-form questions like \textit{``Is the cube above the sphere?''} (v1 answers \texttt{\"false\"} instead of \texttt{"No"}). The 9 structured tasks dominate the training mixture and displace the model's prior for natural-language decoding. Single-view S$^3$-FT (Section~\ref{sec:s3ft:single-view}) addresses this by adding rule-generated NL-QA pairs, scene-description paragraphs, and paraphrased code-gen prompts, restoring the natural-language prior while preserving the structured-task gains.

\subsection{Cross-viewpoint augmentation}
\label{app:s3ft:cross_viewpoint}

\begin{figure}[h]
\centering
\includegraphics[width=0.98\linewidth]{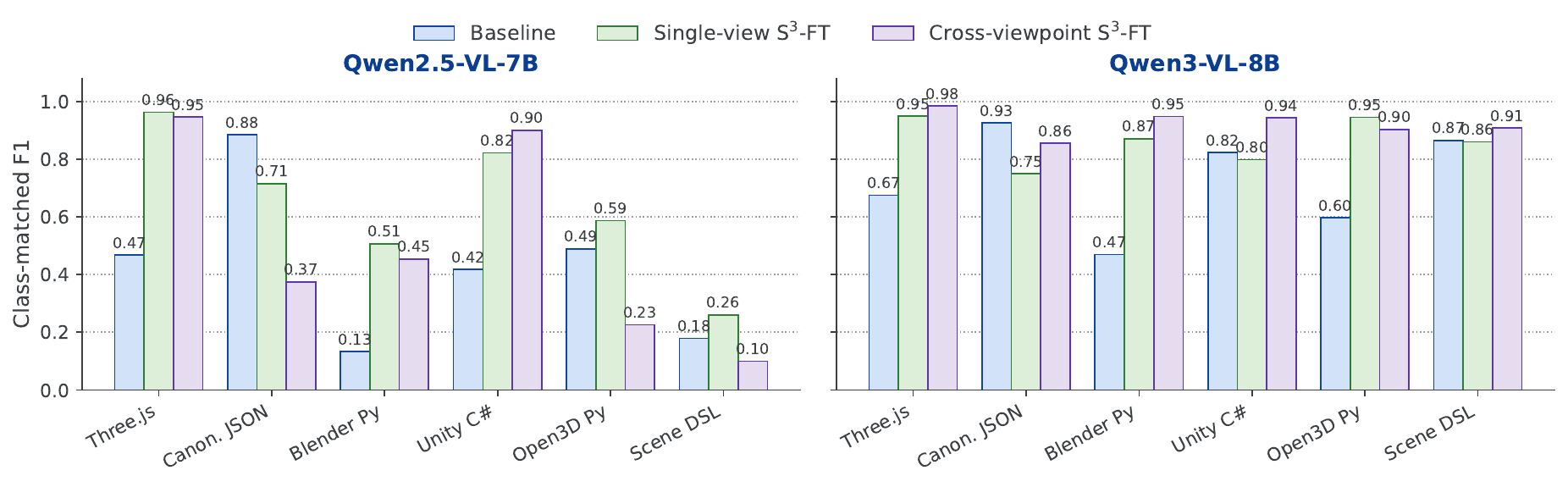}
\caption{Per-language F1 on 100 \emph{primitive} held-out scenes for the two base VLMs, comparing Baseline, v1 (collapsed), single-view S$^3$-FT (our self-supervised recipe) and cross-viewpoint S$^3$-FT.}
\label{fig:s3ft_bench}
\end{figure}

Several of our Phase-2 tasks (object count, classes, scene-coordinate 3D positions, scene description, NL-QA) have answers that do not depend on camera pose, so each such record can be replicated across the other three rendered viewpoints of the same scene without changing the supervision target. We call the resulting $\sim$3.5$\times$ larger dataset \textbf{Cross-viewpoint S$^3$-FT}. This is a supervised analogue of a representation-level cross-view alignment objective: the model is shown that the same scene-coordinate answer must be produced regardless of which 2D projection it sees. Cross-viewpoint S$^3$-FT is the overall winner on Qwen3-VL-8B (mean benchmark F1 $\mathbf{0.924}$, $+0.20$ over baseline), but on the weaker Qwen2.5-VL-7B base it destabilises training (canonical JSON F1 drops from $0.88$ baseline to $0.37$, with only $39/100$ outputs yielding $\geq 1$ parsed object). For the stronger base the extra cross-view signal pays off; for the weaker one Single-view is the safer choice. We adopt Single-view as the headline recipe in the main body for that reason; cross-viewpoint numbers appear alongside in Figure~\ref{fig:s3ft_summary}, Table~\ref{tab:general_vl}, and Figure~\ref{fig:s3ft_bench}.

\subsection{Per-language S\texorpdfstring{$^3$}{3}-FT results}
\label{app:s3ft:per_lang}

The largest absolute gains show up on languages where the base model has the largest output-format margin to close. For Qwen3-VL, Blender Python goes from $0.470$ baseline to $0.948$ cross-viewpoint ($+0.48$), and Three.js from $0.675$ to $0.984$ ($+0.31$). On canonical JSON the base model is already strong ($0.926$) once a tolerant parser is used, so cross-viewpoint S$^3$-FT only matches it ($0.855$); the win there is schema-conformance robustness, not raw F1. S$^3$-FT is not a universal win: on Qwen2.5-VL's Scene Language DSL, neither v1 nor cross-viewpoint improves over baseline; single-view recovers only to $0.260$ (from $0.179$). Across the six languages, single-view is \emph{never} worse than baseline for the stronger base and is worse only on canonical JSON ($0.884 \to 0.715$) for the weaker base.

\subsection{Per-category breakdown of SpatialBabel-QA}
\label{app:s3ft:per_category}

Aggregate SpatialBabel-QA accuracy (Table~\ref{tab:code_cot_sb}) hides where S$^3$-FT actually helps. Breaking the same 800 questions per cell down by category reveals a strikingly localised effect.

\begin{table}[h]
\centering\small
\caption{Per-question-category accuracy (\%) for Qwen3-VL-8B base vs single-view S$^3$-FT on SpatialBabel-QA (100 scenes $\times$ 8 questions = 800 questions per cell, distributed across the 5 categories). $\Delta$ is in percentage points. Bold marks rows where S$^3$-FT improves by more than 5 percentage points.}
\label{tab:s3ft_per_cat}
\resizebox{\linewidth}{!}{%
\begin{tabular}{lccccccccc}
\toprule
& \multicolumn{3}{c}{direct} & \multicolumn{3}{c}{nl\_cot} & \multicolumn{3}{c}{code\_cot (threejs)} \\
\cmidrule(lr){2-4}\cmidrule(lr){5-7}\cmidrule(lr){8-10}
Category     & base & single-view & $\Delta$ & base & single-view & $\Delta$ & base & single-view & $\Delta$ \\
\midrule
localization & 0.0  & 2.0  & \deltaup{2.0}          & 0.7  & 2.0  & \deltaup{1.3}          & 3.4  & 0.0  & \deltadn{3.4} \\
relationship & 35.7 & 41.7 & \textbf{\deltaup{6.0}} & 29.0 & 43.0 & \textbf{\deltaup{14.0}} & 35.0 & 44.9 & \textbf{\deltaup{9.9}} \\
counting     & 72.0 & 79.3 & \textbf{\deltaup{7.3}} & 61.0 & 79.3 & \textbf{\deltaup{18.3}} & 74.4 & 76.8 & \deltaup{2.4} \\
existence    & 97.9 & 92.2 & \deltadn{5.7}          & 88.7 & 91.5 & \deltaup{2.8} & 83.0 & 80.9 & \deltadn{2.1} \\
comparison   & 3.5  & 41.7 & \textbf{\deltaup{38.2}} & 36.5 & 40.0 & \deltaup{3.5}         & 12.2 & 39.1 & \textbf{\deltaup{26.9}} \\
\bottomrule
\end{tabular}}
\end{table}

Three observations from the breakdown.
(i) The aggregate ``single-view S$^3$-FT lifts SpatialBabel-QA by $+7.0$\% on Qwen3-VL-8B'' headline number is dominated by a single category: \textbf{comparison} questions go from $3.5$\% (essentially chance) to $41.7$\%, a $+38$\% lift on direct answering and $+27$\% on code-CoT.
(ii) NL-CoT \emph{counting} ($61.0$\% $\to 79.3$\%, $+18.3$\%) and \emph{relationship} ($29.0$\% $\to 43.0$\%, $+14.0$\%) improve substantially, suggesting structured \texttt{object\_count} and spatial-relation supervision in Phase 2 transfers to free-form NL reasoning under CoT prompting.
(iii) \textbf{Localisation} (precise (x,y,z) within tolerance) is intractable for all configurations, attributable to the small target tolerance ($\sim$0.1 scene units) combined with the VLM's discrete token vocabulary.

\paragraph{Qwen2.5-VL-7B follows the same direct/cc pattern at smaller magnitude.} \emph{Comparison under direct} lifts $\sim$31\% $\to \sim$41\% ($+10$\%, smaller because the Qwen2.5-VL-7B baseline already sits well above chance on this category); \emph{counting under direct} stays near $79$\% across base and S$^3$-FT; \emph{localisation} stays near zero across all configurations. Qwen2.5-VL-7B's NL-CoT improves modestly ($+3.1$\% overall) under the CoT-symmetric evaluator, smaller than Qwen3-VL-8B's $+8.6$\% lift, consistent with Qwen2.5-VL-7B's lower-quality base reconstructions.

\subsection{Multi-seed reliability}
\label{app:s3ft:seeds}

We re-train Qwen3-VL-8B single-view with two additional random seeds (42, 1337) under otherwise identical hyperparameters and compare to the seed-0 checkpoint reported in Table~\ref{tab:code_cot_sb_sft}. All three runs converge to similar val losses ($0.215$ to $0.243$).

\begin{table}[h]
\centering\small
\caption{Single-view S$^3$-FT on Qwen3-VL-8B across multiple independent random seeds. \emph{direct}, \emph{nl\_cot}, \emph{best\_code\_cot (threejs)} are 3-seed runs (0, 42, 1337); \emph{worst\_code\_cot (canonical\_json)} is a 2-seed run (the seed-0 checkpoint was trained with an older configuration not directly comparable on the longest-output mode).}
\label{tab:s3ft_multiseed}
\resizebox{\linewidth}{!}{%
\begin{tabular}{lcccc}
\toprule
Mode & base (\%) & single-view S$^3$-FT seed mean (\%) ($n$) & seed std (pp) & $\Delta$ over base (pp) \\
\midrule
direct                              & 39.1 & 47.4 (3) & 0.4 & \deltaup{8.3 \pm 0.4} \\
nl\_cot                             & 38.6 & 47.3 (3) & 1.5 & \deltaup{8.7 \pm 1.5} \\
best\_code\_cot (threejs)           & 38.4 & 45.9 (3) & 1.0 & \deltaup{7.5 \pm 1.0} \\
worst\_code\_cot (canonical\_json)  & 37.5 & 42.6 (2) & 0.8 & \deltaup{5.1 \pm 0.8} \\
\bottomrule
\end{tabular}}
\end{table}

Across-seed standard deviation is small on every reported mode ($\leq 1.5$\%), giving tight 95\% CIs on the headline lifts: direct $+8.3 \pm 0.5$\%, nl\_cot $+8.7 \pm 1.7$\%, threejs $+7.5 \pm 1.1$\%, canonical\_json $+5.1 \pm 1.1$\%. Under the CoT-symmetric evaluator (Section~\ref{sec:code_cot:setup}) the headline becomes ``single-view S$^3$-FT lifts SpatialBabel-QA by $\sim$+7\% averaged across modes on Qwen3-VL-8B'', with all four modes lifting by $+5$ to $+9$\%.

\subsection{IFEval (text-only) performance comparison}
\label{app:s3ft:ifeval}

To confirm S$^3$-FT does not trade spatial ability for general instruction-following, we run the models on IFEval~\citep{ifeval2023}: 541 prompts with programmatic verifiers for format, length, keyword and language constraints. Prompts are text-only.

\begin{center}
\small
\begin{tabular}{lcccc}
\toprule
Base model & Baseline & S$^3$-FT v1 & single-view S$^3$-FT & cross-viewpoint S$^3$-FT \\
\midrule
Qwen2.5-VL-7B (prompt) & 62.7\% & 61.0\%\,\deltadn{1.7} & 57.3\%\,\deltadn{5.4} & 56.4\%\,\deltadn{6.3} \\
Qwen2.5-VL-7B (instr.) & 71.3\% & 68.5\%\,\deltadn{2.8} & 67.6\%\,\deltadn{3.7} & 66.2\%\,\deltadn{5.1} \\
Qwen3-VL-8B (prompt)   & 83.5\% & \textbf{84.8\%}\,\deltaup{1.3} & 83.0\%\,\deltadn{0.5} & 82.1\%\,\deltadn{1.4} \\
Qwen3-VL-8B (instr.)   & 89.0\% & \textbf{89.7\%}\,\deltaup{0.7} & 88.5\%\,\deltadn{0.5} & 87.8\%\,\deltadn{1.2} \\
\bottomrule
\end{tabular}
\end{center}

For the stronger base (Qwen3-VL-8B), S$^3$-FT is essentially free on IFEval. For the weaker base (Qwen2.5-VL-7B) the cost is larger ($5$ to $6$\% on single-view/cross-viewpoint) because the natural-language supervision targets are structured-but-not-quite-English and dampen the model's prior for long free-form prose. The IFEval delta is small relative to the spatial-F1 gain: Qwen2.5 single-view loses 5.4\% on IFEval but gains 21.4\% on benchmark F1.

\subsection{Cross-base and cross-family validation}
\label{app:s3ft:cross_base}

To test how much of the recipe carries over to (i) a smaller capacity and (ii) different model families, we re-run the full pipeline on \textbf{Qwen2.5-VL-3B-Instruct}, \textbf{InternVL3-8B}, and \textbf{LLaVA-OV-7B}.

\begin{table}[h]
\centering\small
\setlength{\tabcolsep}{4pt}
\caption{Single-view S$^3$-FT across five base VLMs: two headline bases (Qwen3-VL-8B, Qwen2.5-VL-7B) and three cross-family / cross-scale validation bases. Each cell is ``base $\to$ single-view ($\Delta$)'' on programmatic SpatialBabel-QA (100 \emph{primitive} held-out scenes $\times$ 8 questions per cell unless flagged), reported as accuracy in \%; $\Delta$ is in percentage points. The \texttt{cc(blender\_py)} column is reported for the bottom three cross-base rows because Blender Python is each of those models' primitive-best language (Qwen2.5-VL-3B .667, InternVL3-8B .649, LLaVA-OV-7B .648; Table~\ref{tab:main_results}); for the two headline bases at the top, Blender Python is each model's primitive-worst (Qwen3-VL-8B .514, Qwen2.5-VL-7B .320) and the post-S$^3$-FT × Blender Python evaluation was deprioritized in favour of the primitive-best/worst runs reported in Table~\ref{tab:code_cot_sb_sft}.}
\label{tab:s3ft_cross_base}
\resizebox{\linewidth}{!}{%
\begin{tabular}{lccccc}
\toprule
Base VLM & direct & nl\_cot & cc(threejs) & cc(canonical\_json) & cc(blender\_py) \\
\midrule
Qwen3-VL-8B    & $39.1\!\to\!47.1$\deltaup{8.0} & $38.6\!\to\!47.2$\deltaup{8.6} & $38.4\!\to\!45.4$\deltaup{7.0} & $37.5\!\to\!42.1$\deltaup{4.6} & --- \\
Qwen2.5-VL-7B  & $45.4\!\to\!48.6$\deltaup{3.2} & $44.4\!\to\!47.5$\deltaup{3.1} & $44.5\!\to\!46.2$\deltaup{1.7} & $47.1\!\to\!38.9$\deltadn{8.2} & --- \\
\midrule
Qwen2.5-VL-3B  & $36.4\!\to\!39.1$\deltaup{2.7} & $41.8\!\to\!38.9$\deltadn{2.9} & $37.2\!\to\!37.8$\deltaup{0.6} & $23.4\!\to\!33.4$\deltaup{10.0} & $20.2^\star\!\to\!39.5$\deltaup{19.3} \\
InternVL3-8B   & $37.5\!\to\!43.5$\deltaup{6.0} & $41.4\!\to\!44.0$\deltaup{2.6} & $40.5\!\to\!34.1$\deltadn{6.4} & $46.9\!\to\!41.8$\deltadn{5.1} & $37.9\!\to\!43.1$\deltaup{5.2} \\
LLaVA-OV-7B    & $40.0\!\to\!43.4$\deltaup{3.4} & $37.1\!\to\!43.4$\deltaup{6.3} & $46.0\!\to\!42.1$\deltadn{3.9} & $49.4\!\to\!43.0$\deltadn{6.4} & $48.2\!\to\!42.5$\deltadn{5.7} \\
\bottomrule
\end{tabular}}
\\\smallskip
{\footnotesize $^\star$ Qwen2.5-VL-3B base blender\_py is over 50 scenes (n=400 questions) due to verbose unterminated output. Its single-view S$^3$-FT counterpart (which terminates output cleanly) is over 100 scenes.}
\end{table}

\textbf{Three patterns.} (i) \textbf{Direct mode lifts uniformly across bases.} Every base in the table gains on direct: Qwen3-VL-8B $+8.0$, InternVL3-8B $+6.0$, LLaVA-OV-7B $+3.4$, Qwen2.5-VL-7B $+3.2$, Qwen2.5-VL-3B $+2.7$. This is the cleanest S$^3$-FT signal in the cross-base sweep. (ii) \textbf{Code-CoT lifts the base's weakest languages most.} On Qwen2.5-VL-3B the largest cc gains land on its primitive-worst languages: canonical\_json $+10.0$\% and blender\_py $+19.3$\% (the two cells where the base scored 23\% to 28\%). Where the recipe regresses cc cells (LLaVA-OV-7B and InternVL3-8B), the regressions cluster on the base's primitive-best language, consistent with same-distribution distillation overwriting the model's strongest CoT habits. (iii) \textbf{NL-CoT lifts only when the base CoT is poorly structured.} LLaVA-OV-7B's NL-CoT lifts $+6.3$\% (the base produces unstructured prose that benefits from S$^3$-FT's ``Final answer:'' formatting), whereas Qwen3-VL-8B / Qwen2.5-VL-7B / Qwen2.5-VL-3B / InternVL3-8B regress on NL-CoT under the CoT-tolerant evaluator: the bases were already producing committed-answer text that scored fairly under the new parser. The recipe transfers across model architectures: LLaVA-OV-7B uses a different visual encoder and patch-merger from Qwen-VL or InternVL, yet the same pipeline lifts its direct-mode SpatialBabel-QA without changes.

\subsection{Hypersim: when the teacher is not good enough}
\label{app:s3ft:hypersim}

We rebuild the full single-view S$^3$-FT pipeline on SpatialBabel's 200 Hypersim indoor scenes: Phase-1 pseudo-GT generation with the same two base VLMs, Phase-2 annotation extraction, Phase-3 LoRA fine-tuning on 180 scenes with 20 held out for evaluation.

\begin{table}[h]
\centering\small
\caption{Single-view S$^3$-FT on Hypersim (20 held-out scenes $\times$ 8 questions = 160 per cell), reported as accuracy in \%; $\Delta$ is in percentage points. Under the CoT-symmetric evaluator, S$^3$-FT lifts direct and NL-CoT modes for both Qwen models on Hypersim, but Code-CoT modes mostly regress; Qwen3-VL-8B code-CoT canonical\_json collapses ($23.6$\% $\to 3.2$\%) due to the model emitting near-empty Three.js code under the rare canonical\_json prompt format that does not appear in the S$^3$-FT training distribution.}
\label{tab:s3ft_hypersim}
\begin{tabular}{lccc}
\toprule
\textbf{Qwen3-VL-8B} (Hypersim)  & Base & single-view S$^3$-FT & $\Delta$ \\
\midrule
direct                         & 25.5 & 28.0          & \deltaup{2.5} \\
nl\_cot                        & 21.7 & 27.4          & \deltaup{5.7} \\
code\_cot (threejs)            & 22.3 & 21.7          & \deltadn{0.6} \\
code\_cot (canonical\_json)    & 23.6 & \textit{3.2}  & \deltadn{20.4} \\
\midrule
\textbf{Qwen2.5-VL-7B} (Hypersim) & Base & single-view S$^3$-FT & $\Delta$ \\
\midrule
direct                         & 28.0 & 28.0 & \deltazero{0.0} \\
nl\_cot                        & 22.9 & 29.3 & \deltaup{6.4} \\
code\_cot (threejs)            & 30.6 & 22.3 & \deltadn{8.3} \\
code\_cot (blender\_py)        & 27.4 & 25.5 & \deltadn{1.9} \\
\bottomrule
\end{tabular}
\end{table}

\textbf{Why the recipe fails here.} The training pipeline is mechanically identical to the primitive one; the pseudo-GT pass rate is also comparable (68\% to 58\%). The breaking point is the \emph{fidelity} of the pseudo-GT code to the actual scene content. The pseudo-GT Three.js produced by the base VLMs averages $\sim$9 to 10 objects per Hypersim scene, whereas Hypersim's ground-truth scene graphs have a median of 26 objects per scene (mean 66, IQR 11 to 58). A Phase-1 ``reconstruction'' that captures less than 1/3 of the objects present no longer qualifies as a pseudo-GT: every downstream task is trained against a systematically incomplete label.

This is a qualitative phase transition. On SpatialBabel primitives, pseudo-GT covers $\sim$89\% of scene content (8.4 vs 9.4 objects/scene) and the recipe produces large gains. On Hypersim coverage is only $\sim$38\% (9.8 vs 26) and the same recipe yields no gain or regresses. We conclude S$^3$-FT's headline result is \emph{conditional on the teacher VLM's code-generation ability spanning the target domain's scene complexity}; beyond that threshold, the ``oracle'' in ``pseudo-oracle'' is no longer an oracle.

Two paths forward: (i) a stronger teacher at the cost of self-supervision; (ii) FOV-aware target derivation at training time so Phase-2 tasks only cover visible objects, which raises pseudo-GT coverage to $\sim$0.16 of the in-frustum portion (still low, but cleaner training signal). We list both as follow-ups.

\section{Broader Impacts}
\label{app:broader_impacts}

Stronger spatial understanding in VLMs is a foundational capability for embodied agents (robotic manipulation, autonomous navigation), assistive technology (scene description for blind and low-vision users), AR/VR scene authoring, and educational tools that explain 3D structure from a single image. The self-supervised nature of S$^3$-FT (which requires \emph{no human spatial labels, depth sensors, or frontier-teacher distillation}) lowers the cost and labour barrier for fine-tuning open-weights VLMs, broadening access for academic groups and small organizations that cannot afford large-scale annotation pipelines or proprietary teachers. Releasing the benchmark, training data, parser toolkit, and LoRA adapters under permissive licenses upon publication further democratizes spatial-VLM research. Potential misuses (e.g., surveillance) are mitigated by the benchmark containing no human imagery and the LoRA-only release allowing downstream users to opt-out by simply not loading the adapter.

%% file: sections/appendix_reproducibility.tex
\section{Reproducibility appendix}
\label{app:reproducibility}

We include the following to support reproduction of every numeric result in the paper.

\subsection{S$^3$-FT hyperparameters}

\begin{center}
\small
\begin{tabular}{ll}
\toprule
\textbf{Hyperparameter} & \textbf{Value} \\
\midrule
Base models           & \texttt{Qwen/Qwen2.5-VL-7B-Instruct}, \texttt{Qwen/Qwen3-VL-8B-Instruct} \\
LoRA                  & rank 16, $\alpha = 32$, dropout 0.05, target modules \{q,k,v,o\}\_proj \\
Optimizer             & AdamW, lr $2\mathrm{e}{-4}$, weight decay 0 \\
Schedule              & cosine decay, no warmup \\
Batch / accumulation  & batch size 1, gradient accumulation 8 (effective 8) \\
Epochs                & 3 for v1 (deprecated) and single-view S$^3$-FT; 2 for cross-viewpoint (larger data) \\
Sequence length       & 2048 tokens max \\
Mixed precision       & bfloat16 \\
Checkpoint selection  & best validation loss over epochs; final model saved additionally \\
\bottomrule
\end{tabular}
\end{center}

\subsection{Phase-1 quality filters}

Generated code is accepted only if it contains all three tokens \{\texttt{THREE.Scene}, \texttt{THREE.PerspectiveCamera}, \texttt{THREE.Mesh}\}, has 1 to 50 \texttt{new THREE.Mesh(} occurrences, and all \texttt{position.set(x,y,z)} values lie in $[-30, 30]$. Pass rates: Qwen3-VL-8B 82\%, Qwen2.5-VL-7B 84\%.

\subsection{Phase-2 task mixture}

single-view S$^3$-FT (per scene): 9 JSON task records + 1 scene description (free-form) + 8 NL-QA pairs (localisation, relationship, counting, existence, comparison via rule-based generator) + 2 extra code-generation prompt paraphrases. Dataset size after Phase-1 filtering: $\sim$3.5k records per model. Cross-viewpoint S$^3$-FT replicates every view-invariant record across the four rendered viewpoints, $\sim$3.5$\times$ data growth.

\subsection{SpatialBabel eval protocol}

The test set is 100 held-out primitive scenes (pre-filtered from the Phase-1 candidate pool to avoid leakage). Generation uses temperature 0.0, \texttt{max\_new\_tokens} 2048, and no beam search. Scoring is parser-based: each generated code string is parsed into a \texttt{Scene} object, Hungarian-matched against the ground truth, and graded with class-matched F1 at IoU threshold 0.5. Confidence intervals are 95\% bootstrap over scenes with 1{,}000 resamples.

\subsection{Public VLM benchmark protocol}

VLMEvalKit, generation-based evaluation with \texttt{use\_custom\_prompt=True} (standard MCQ template) for every configuration. Baseline models use the same wrapper; S$^3$-FT adapters are loaded via PEFT at model-init time. Fallback to \texttt{sdpa} attention when flash\_attn is unavailable; we verified no numeric difference on POPE / OCRBench. Full per-benchmark numbers in \texttt{results/vlmeval/final\_summary.json} in the code release.

\subsection{Software versions}

transformers $\geq$4.46, peft $\geq$0.13, torch 2.x, VLMEvalKit main@2026-04. numpy pinned to \textless2 for compatibility with downstream pandas used in VLMEvalKit. IFEval scoring via the \texttt{lm\_eval.tasks.leaderboard.ifeval} module (official Google-Research port).

\subsection{Data splits}

Primitive scenes are split deterministically: 1{,}000 total, 101 held out for evaluation (stratified across tiers 1 to 5), 467 used for Phase-1 pseudo-GT generation (the remaining $\sim$430 are kept unused as a safety margin). Train / val for Phase-3 LoRA is a further 90/10 split of the Phase-1 pseudo-GT survivors (seed 42), grouped by scene so no scene appears in both.

We commit two aggregated snapshot files: \texttt{results/benchmark/aggregate.json} (the wide table read by every figure script) and \texttt{results/vlmeval/final\_summary.json} (the wide table read by Table~\ref{tab:general_vl}). These contain the per-cell aggregate metrics needed to regenerate every table and figure in the paper.

\paragraph{Availability.} Upon publication, all artefacts will be uploaded to a public Zenodo record (DOI minted at submission time) and a mirror GitHub repository. The release will include: (i) the aggregated per-cell metric snapshots used to regenerate every table and figure, (ii) checkpoint files for all S$^3$-FT adapters, (iii) the parser/compiler suite as an installable package, and (iv) Docker images for the evaluation harness so reviewers can rerun any benchmark cell without provisioning the upstream VLMEvalKit environment.